\documentclass[lettersize,journal]{IEEEtran}
\usepackage{amsfonts}

\usepackage{array}
\usepackage[caption=false,font=normalsize,labelfont=sf,textfont=sf]{subfig}
\usepackage{textcomp}
\usepackage{stfloats}
\usepackage{url}
\usepackage{verbatim}
\usepackage{multicol}
\usepackage{multirow}
\usepackage{threeparttable}
\usepackage{booktabs}
\usepackage{colortbl,xcolor,array}

\usepackage{algpseudocode}

\usepackage{algorithm}  
\usepackage{algorithmicx}

\usepackage{newtxmath}
\usepackage{graphicx}
\usepackage{cite}
\begin{document}

\title{Multi-Timescale Motion-Decoupled Spiking Transformer for Audio-Visual Zero-Shot Learning}

\author{Wenrui Li,
        Penghong Wang,
        Xingtao Wang,
        % Xi-Le Zhao,~\IEEEmembership{Member,~IEEE}
        Wangmeng Zuo,~\IEEEmembership{~Senior Member,~IEEE}\\
        Xiaopeng Fan,~\IEEEmembership{~Senior Member,~IEEE}
        Yonghong Tian,~\IEEEmembership{~Fellow,~IEEE}

\thanks{This work was supported in part by the National Key R\&D Program of China (2023YFA1008500), the National Natural Science Foundation of China (NSFC) under grants 624B2049, 62402138, and the Fundamental Research Funds for the Central Universities under grants HIT.DZJJ.2024025. (Corresponding author: Xiaopeng Fan.)}
\thanks{Wenrui Li and Wangmeng Zuo are with the Department of Computer Science and Technology, Harbin Institute of Technology, Harbin 150001, China. (e-mail: liwr@stu.hit.edu.cn; wmzuo@hit.edu.cn).}
\thanks{Penghong Wang, Xingtao Wang and Xiaopeng Fan are with the Department of Computer Science and Technology, Harbin Institute of Technology, Harbin 150001, China, and also with Harbin Institute of Technology Suzhou Research Institute, Suzhou 215104, China. (e-mail:  phwang@hit.edu.cn; xtwang@hit.edu.cn; fxp@hit.edu.cn).}
\thanks{Yonghong Tian is with the School of AI for Science, the Shenzhen Graduate School, Peking University, Shenzhen, China, the Peng Cheng Laboratory, Shenzhen, China and also with the School of Computer Science, Peking University, Beijing, China (e-mail: yhtian@pku.edu.cn).}
% \thanks{Ruiqin Xiong is with the School of Electronic Engineering and Computer Science, Institute of Digital Media, Peking University, Beijing 100871, China (e-mail: rqxiong@pku.edu.cn).}
}

% The paper headers
% \markboth{Journal of \LaTeX\ Class Files,~Vol.~14, No.~8, August~2021}%
% {Shell \MakeLowercase{\textit{et al.}}: A Sample Article Using IEEEtran.cls for IEEE Journals}

% Remember, if you use this you must call \IEEEpubidadjcol in the second
% column for its text to clear the IEEEpubid mark.

\maketitle

\begin{abstract}
Audio-visual zero-shot learning (ZSL) has been extensively researched for its capability to classify video data from unseen classes during training. Nevertheless, current methodologies often struggle with background scene biases and inadequate motion detail. This paper proposes a novel dual-stream Multi-Timescale Motion-Decoupled Spiking Transformer (MDST++), which decouples contextual semantic information and sparse dynamic motion information. The recurrent joint learning unit is proposed to extract contextual semantic information and capture joint knowledge across various modalities to understand the environment of actions. By converting RGB images to events, our method captures motion information more accurately and mitigates background scene biases. Moreover, we introduce a discrepancy analysis block to model audio motion information. To enhance the robustness of SNNs in extracting temporal and motion cues, we dynamically adjust the threshold of Leaky Integrate-and-Fire neurons based on global motion and contextual semantic information. Our experiments validate the effectiveness of MDST++, demonstrating their consistent superiority over state-of-the-art methods on mainstream benchmarks. Additionally, incorporating motion and multi-timescale information significantly improves HM and ZSL accuracy by 26.2\% and 39.9\%.
\end{abstract}
 % MDST++ utilizes spiking self-attention with a multi-stage spiking timestep shrinkage strategy to further explore the long-range dependencies and multi-scale time information to optimize parameter updates.
\section{Introduction}
\IEEEPARstart{Z}{ero}-Shot learning (ZSL) tasks in the audio-visual domain aim to use both audio and visual modalities to classify target classes that have not been observed during training. Supervised audio-visual methods \cite{cheng2,cheng3,ouyang2,zhanglei2} train models with input data and corresponding labels, allowing them to predict new inputs. However, previous methods are limited in recognizing previously unseen classes, restricting them to categorizing only the classes in the training data. Training models for every possible data class in real-world settings is impractical. Therefore, generalized zero-shot learning extends ZSL by aiming to classify both unseen and seen classes during testing. Currently, most of the existing audio-visual ZSL methods \cite{EC46mazumder2021avgzslnet, EC47mercea2022audio,TCAF22} have focused on projecting temporal and semantic information accurately to generate more precise audio-visual representations. However, these methods mainly focus on effectively modeling and integrating temporal and semantic information, overlooking two critical factors: \textbf{background scene bias} and \textbf{motion information}.

Previous research \cite{wang2021enhancing,Li-tip24} has demonstrated that specific action categories in mainstream video datasets are closely related to the background scenes where the actions occur. When models overemphasize contextual semantics, they will degrade by primarily encoding fine-grained scene details. For instance, a model might misclassify a video as ``Running" simply because a running track is presented, even if the actual activity is ``Baby Sleeping". This issue conflicts with the fundamental goal of video representation learning introduces background scene bias. This is a significant concern when applied to diverse datasets. Decoupling motion information from video inputs to optimize multimodal feature learning is still challenging.
\begin{figure}
	\centering
	\includegraphics[scale=0.5]{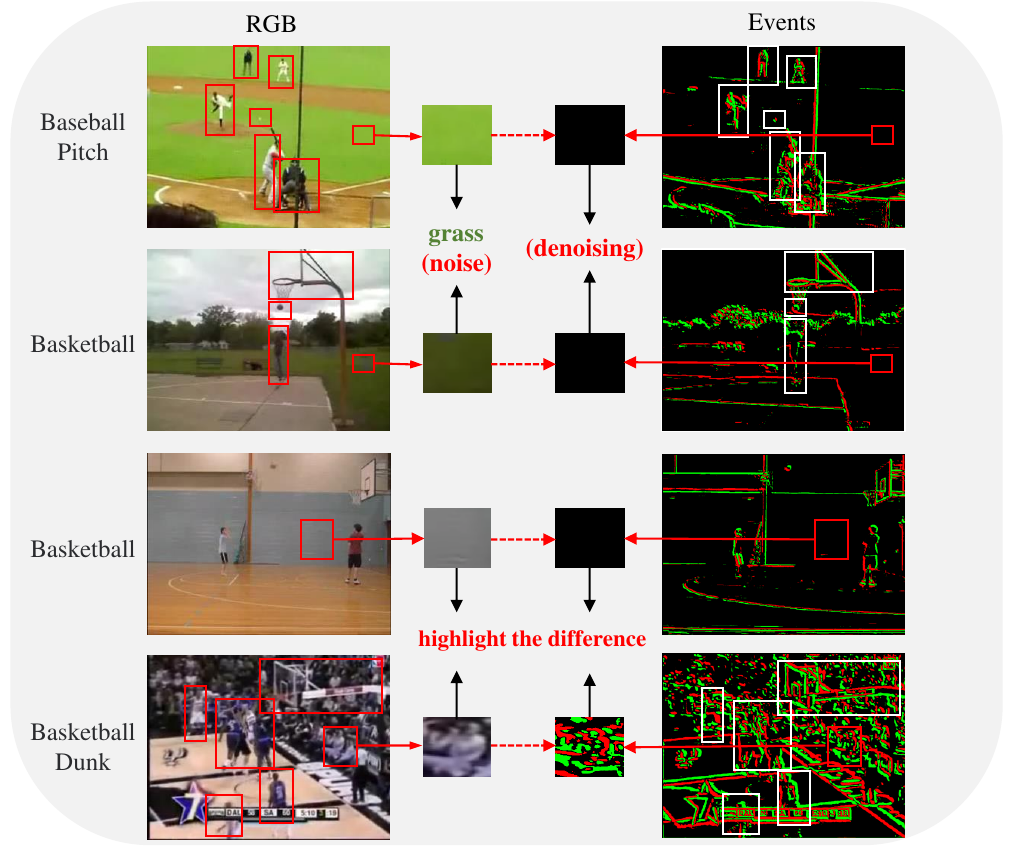}
	\caption{To mitigate background bias and highlight the differences among similar types of videos, we converted RGB images into events. This conversion only occurs when there are significant changes in the background scene.}
	\label{fig:1}
\end{figure}

To address these challenges, we propose an Event Generation Model (EGM) to convert RGB images into events, eliminating background scene bias and capturing essential motion details. As shown in Fig. \ref{fig:1}, our EGM effectively removes background noise from images in scenarios such as ``Baseball" and ``Basketball Dunk". Moreover, EGM captures and highlights critical motion elements, such as the movements of players and objects. For action categories with similar semantic contexts, such as ``Basketball" and ``Basketball Dunk", our method effectively distinguish them by capturing distinct motion information, including audience reactions, player movements, and ball positions. Event data, characterized by the time and location of events, exhibits spatial and temporal sparsity. We utilize Spiking Neural Networks (SNNs) to process sparse event data, which naturally encode temporal information and directly extract features without additional preprocessing or dimensionality reduction. Moreover, the memory capabilities of SNNs make them well-suited for attention mechanisms. When utilizing SNNs to process video data with long contextual dependencies, it is crucial to ensure the following two aspects: 

1) \textbf{Developing Spiking Attention Mechanisms}: Implement spiking attention to enhance the long-term dependencies of spiking features. This integration enables SNNs to effectively capture and interpret complex temporal dynamics in the data, significantly improving performance in tasks requiring long-range dependency understanding.

2) \textbf{Effectively Utilizing Timestep Information}: Wisely use timestep information to reduce spiking redundancy. By employing a multi-stage timestep shrinkage strategy, the SNN maintains efficiency in feature extraction while minimizing inference latency. This approach ensures each network layer captures important features at its specific temporal resolution, optimizing the accuracy and efficiency of processing complex audio-visual data.

This paper aims to separate contextual semantic information from dynamic motion data to minimize background scene bias and enhance the focus on motion for video classification. We use a Recurrent Joint Learning Unit (RJLU) to integrate latent information from various modalities, promoting efficient joint learning. The RJLU helps the network understand contextual semantics and the environments where actions occur. By combining the Event Generation Model (EGM) with Spiking Neural Networks (SNNs), our model effectively captures both motion and temporal information. To tackle the challenge of capturing motion cues from audio, we introduce a Discrepancy analysis block (DAB) that models the differences between converted visual events and extracted audio information, defining these differences as audio motion cues. To enhance the robustness of SNNs in capturing dynamic temporal and motion information, we also dynamically adjust the thresholds of Leaky Integrate-and-Fire (LIF) neurons based on statistical cues from global motion and contextual semantics. To better capture long-range dependencies in SNNs, MDST++ combines SNNs with self-attention mechanisms to further enhance the model's learning capability. To fully exploit multi-scale temporal information, we propose a multi-stage timestep shrinkage strategy that ensures information retention by gradually decreasing time steps. These advancements make MDST++ particularly effective in handling complex sequential and event data patterns, resulting in superior classification performance. The main contributions could be summarised as follows:
\begin{itemize}

\item We introduce an innovative dual-stream architecture MDST to decouple background scene and motion information effectively. This is the first work that utilizes synthetic events with SNNs for audio-visual ZSL.

\item We utilize the EGM and SNNs to capture sparse motion information and mitigate background scene bias. By dynamically adjusting the threshold of spiking neurons, we further enhanced the robustness and efficiency of SNNs in processing temporal and motion information.

\item We propose the RJLU for extracting contextual semantic information. The fusion block within RJLU combines scene and motion data, facilitating more precise and comprehensive inference.

\item We present MDST++, which integrates SNN with self-attention to capture long-range dependencies better. To further reason multi-scale temporal information, we introduce a multi-stage timestep shrinkage strategy that preserves information by gradually reducing time steps.

\end{itemize}

The previous version of this paper was published in ACM MM 2023 \cite{MDST}. This updated version includes: (1) an enhanced motion information modeling algorithm leveraging spiking attention to improve the capture of long-term dependencies (2) a multi-stage timestep shrinkage strategy that preserves information by gradually reducing timesteps to enhance multi-scale temporal reasoning; (3) additional fine-grained visualization of proposed method; (4) an expanded algorithm training framework for both MDST and MDST++; (5) more detailed experiments demonstrating the advantages of our method. We demonstrate the superiority of our method over the current state-of-the-art methods through extensive experimental results. Furthermore, we conducted numerous ablation studies to verify the effectiveness of each key component in MDST. 

% The rest of the paper is organized as follows: Section \ref{relatedwork} provides a comprehensive background for our study, focusing on audio-visual zero-shot learning, synthetic events, and spiking neural networks. In Section \ref{proposedmethod}, we present a detailed description of our proposed MDST and MDST++ architectures. Section \ref{experiment} showcases the experimental results and visualizations that illustrate the efficacy of our model. Finally, Section \ref{conclusion} summarizes the key findings and contributions of our work.
\section{Related work}
\label{relatedwork}
 %  \cite{TCAF22} proposes An architecture consisting of a temporal attention module and a cross-modal attention module, which jointly learns to align and fuse the audio and visual features of the input data. 
\subsection{Audio-Visual Zero-Shot Learning}
Due to the powerful representational capabilities of deep learning \cite{cheng1,ouyang1,zhanglei1,Chen_2023_ICCV,tang1,zhuoyuan1,zhuoyuan2,ZZChen1}, many audio-visual ZSL methods \cite{av1,av2,av3,av4,av5,tcsvt1,tcsvt2,tcsvt3,tcsvt4,tcsvt5,av14} focus on learning a shared embedding space to inference the latent relationships between these two modalities. Avgzslnet \cite{EC46mazumder2021avgzslnet} introduces a novel GZSL approach in a multimodal setting by aligning audio and video embeddings with class-label text features. It employs a cross-modal decoder and composite triplet loss to enhance performance even with missing modalities during testing. ActivityNet \cite{CV26caba2015activitynet} introduces a large-scale video benchmark for human activity recognition. It addresses the limitations of current benchmarks by covering a wide range of complex activities with 203 classes and 849 hours of untrimmed videos, facilitating comparison in untrimmed video classification, trimmed activity classification, and activity detection. VAEGAN \cite{av1} enforces semantic consistency at all stages of GZSL using a feedback loop to refine features iteratively. This improves classification accuracy and outperforms existing methods on six benchmarks. CADA-VAE \cite{av2} proposes using modality-specific aligned variational autoencoders to learn a shared latent space for image features and class embeddings. Unlike previous methods, our approach simplifies this task by decoupling scene and motion information. Integrating the Event Generative Model (EGM) with Spiking Neural Networks (SNNs), our method effectively captures motion and temporal information while minimizing background scene noise, resulting in superior zero-shot classification performance.
\subsection{Synthetic Events}
The event camera \cite{eventcamera1} is an advanced vision sensor that captures changes in a scene with remarkable speed and precision. When a pixel detects a change in light intensity, it generates an event, recording the pixel's location, the time of the change, and the intensity difference. Additionally, synthetic events \cite{events1,events2} can be generated to enhance event cameras' functionality. These synthetic events are created to simulate real events, assisting in tasks like data augmentation, algorithm testing, and performance evaluation without needing actual physical events. This unique mechanism offers several advantages: event cameras can capture extremely fast movements without motion blur, operate with low latency as events are recorded in real time, and function effectively in both very bright and very dark environments due to their high dynamic range. Furthermore, because they only record changes instead of entire frames, event cameras consume significantly less power than traditional cameras. These features make event cameras particularly valuable in applications like robotics \cite{eventrobot1,eventrobot2}, autonomous vehicles \cite{eventsvi1,eventsvi2}, and high-speed video analysis \cite{eventsvideo1,eventsvideo2,e18}. Rebecq et al. \cite{e32} introduced the first event camera simulator capable of generating reliable synthetic event data for research, utilizing an adaptive rendering scheme to efficiently sample frames, aiding in algorithm prototyping, deep learning, and benchmarking. The ability to efficiently process high-frequency data while maintaining accuracy highlights the transformative potential of event cameras in evolving technological landscapes.

\subsection{Spiking Neural Network}
Spiking Neural Networks (SNNs) significantly advance artificial neural networks by closely mimicking the functionality of biological brains. Several studies \cite{s1,s6,spikemba,zhou2022spikformerspikingneuralnetwork,zhao1,zhao2} explore SNN training methodologies, application contexts, and biological resemblances in detail. Unlike traditional neural networks that rely on continuous values and weighted sums, SNNs communicate through discrete events called spikes, similar to action potentials in biological neurons. SNNs operate on the principle that neurons communicate by sending electrical pulses or spikes, with the timing of these spikes crucial for information processing and transmission. The basic functioning of an SNN involves three stages: input encoding, neuronal processing, and output generation. Information is encoded into spikes and fed into the network, where neurons process these spikes based on timing. Neurons generate output spikes when their membrane potential exceeds a threshold, resulting in processed information as a series of output spikes. Among various spiking neuron models, the Leaky Integrate-and-Fire (LIF) neuron has gained prominence due to its balance between biological realism and computational simplicity. The LIF neuron captures the core dynamics of neuronal membrane potential through a differential equation that governs the integration of incoming synaptic currents and the natural leakage of charge over time:
\[
\tau_m \frac{dV(t)}{dt} = -V(t) + RI(t),
\]
where \(V(t)\) is the membrane potential, \(I(t)\) denotes the input current from presynaptic spikes, \(R\) is the membrane resistance, and \(\tau_m\) is the membrane time constant reflecting how quickly the potential decays. This model simulates the capacitive charging and resistive leakage behaviour of biological neurons: as input spikes arrive, the membrane potential accumulates (charging), and in the absence of input, it decays exponentially (leakage). When the accumulated potential \(V(t)\) exceeds a predefined threshold \(V_{th}\), the neuron emits a spike, after which the membrane potential is reset to a lower value \(V_{reset}\), often accompanied by a brief refractory period during which the neuron is unresponsive to further input. Formally, this spiking mechanism is defined as:
\[
\text{If } V(t) \geq V_{th} \rightarrow \text{neuron spikes}, \quad V(t) \leftarrow V_{reset}.
\]

Training an SNN typically involves adjusting synaptic weights based on the correlation between pre- and post-synaptic spikes. A key principle underlying this process is spike-timing-dependent plasticity (STDP), a biologically inspired rule that modifies synaptic strength based on the precise timing difference between spikes of connected neurons. Specifically, if a presynaptic neuron fires shortly before a postsynaptic neuron, the synapse is strengthened (potentiated); if the firing order is reversed, the synapse is weakened. This asymmetric Hebbian rule enables the network to capture temporal causality in data. Synaptic plasticity thus facilitates unsupervised, local learning, enabling the network to adapt to input statistics and dynamically form internal representations. In addition to STDP, recent methods incorporate global learning signals via surrogate gradients, enabling supervised training through backpropagation in SNNs despite the non-differentiable nature of spike events. These methods approximate the gradient of the spiking function, enabling deep SNNs to be trained similarly to ANNs while preserving temporal dynamics and sparse computation. The unique properties of SNNs, including their event-driven architecture, low power consumption, and temporal coding capability, make them particularly suitable for a range of temporal and sensory-rich applications. These applications include pattern recognition \cite{snnpr1,snnpr2,NSTRN,Li-tmm24}, where temporal spike patterns represent complex spatial features; video processing \cite{li2023modality,MDST}, where frame sequences naturally translate into time-dependent spike trains; and robotics \cite{snnro1,snnro2}, where real-time decision-making and energy efficiency are essential for embedded systems and neuromorphic control. Ongoing research on SNNs continues to explore their potential to create biologically plausible and computationally efficient AI systems, bridging the gap between neuroscience and machine learning.

\begin{figure*}
	\centering
	\includegraphics[scale=0.58]{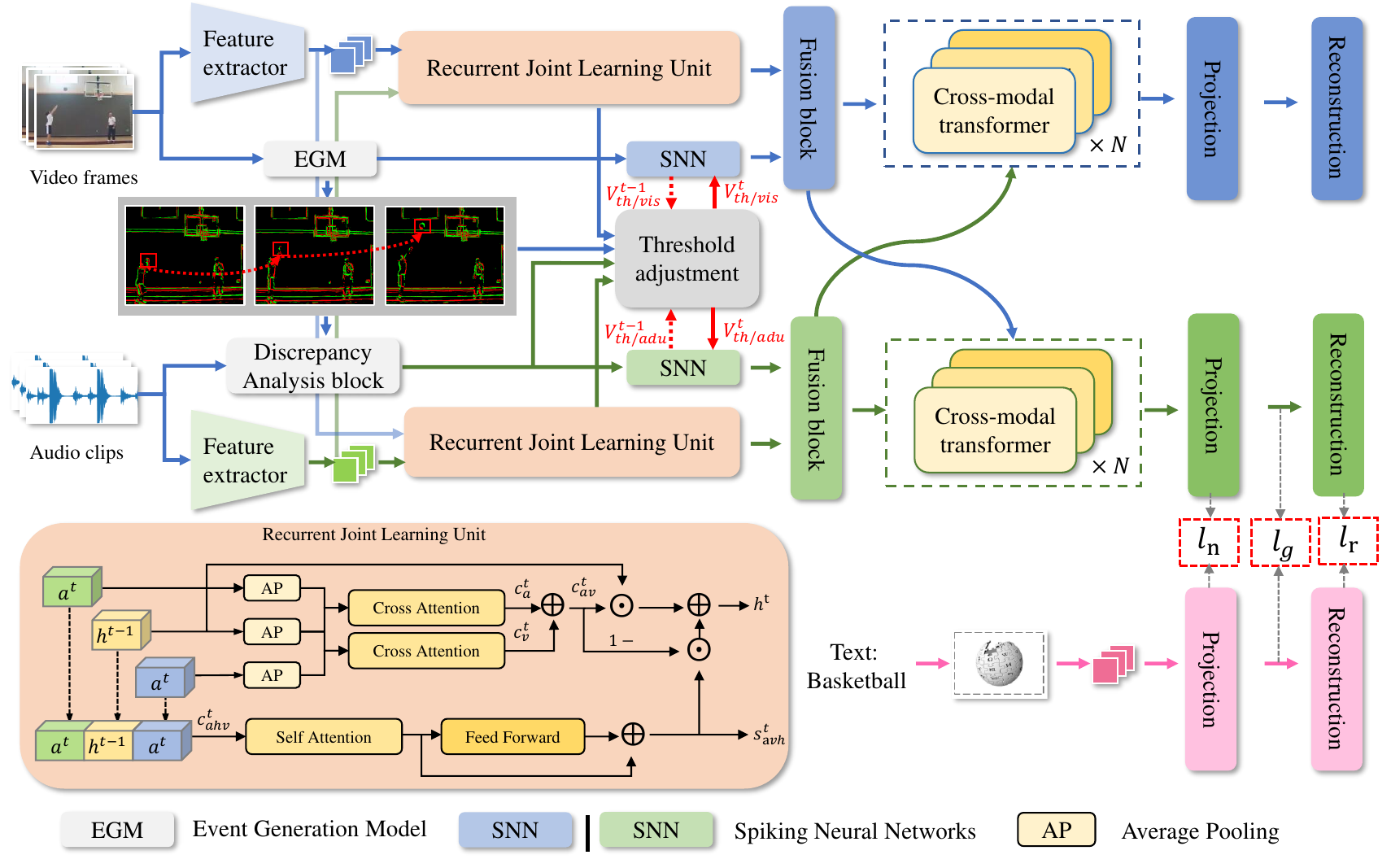}
% 	[height=8cm,width=18cm] [scale=0.78]
	\caption{The MDST architecture combines visual, audio, and textual features (represented by blue, green, and pink lines, respectively) using a two-stream design to extract scene contextual semantics and motion information separately. The ``threshold adjustment" block dynamically modifies the SNN thresholds ($V_{th/vis}^{t}$ and $V_{th/adu}^{t}$) to regulate neuron firing rates and reduce potential noise efficiently.}
	\label{fig:2}
\end{figure*}

\section{Multi-Timescale Motion-Decoupled Spiking Transformer (MDST++)} \label{proposedmethod}
Audio and visual features are represented as $\boldsymbol{a}_{i}^{x}$ and $\boldsymbol{v}_{i}^{x}$, respectively, while the textual labeled embedding of the corresponding ground-truth class $i$ is denoted as $\boldsymbol{t}_{i}^{x}$. During training, the training set for seen classes with $N$ samples is denoted as $\boldsymbol{\mathcal{X}} = (\boldsymbol{a}_{i}^{x}, \boldsymbol{v}_{i}^{x}, \boldsymbol{t}_{i}^{x})$. The Motion-Decoupled Spiking Transformer \cite{MDST} (MDST) is designed to learn a projection function: $ f(\boldsymbol{a}_{i}^{y}, \boldsymbol{v}_{i}^{y}) \mapsto \boldsymbol{g}_{j}^{y} $, where $\boldsymbol{g}_{j}^{x}$ is the class-level textual embedding for class $j$. The unseen testing set $\boldsymbol{\mathcal{Y}} = (\boldsymbol{a}_{i}^{y}, \boldsymbol{v}_{i}^{y}, \boldsymbol{t}_{i}^{y})$ can also be projected using the same function: $ f(\boldsymbol{a}_{i}^{y}, \boldsymbol{v}_{i}^{y}) \mapsto \boldsymbol{g}_{j}^{y}$. The architecture of MDST++, illustrated in Fig. \ref{fig:2}, is specifically designed to handle both contextual semantic information and dynamic motion information separately. This design effectively decouples scene and motion information, enabling better performance in zero-shot learning scenarios.

In summary, the MDST++ begins by extracting high-level audio and visual embeddings through modality-specific encoders. These embeddings are fed into the Recurrent Joint Learning Unit (RJLU), which performs recurrent fusion to generate temporally coherent joint semantics. Simultaneously, the Event Generation Model (EGM) converts raw RGB frames into sparse event streams, allowing the model to suppress background noise and highlight motion cues. These event streams are then processed by Spiking Neural Networks (SNNs) to encode fine-grained temporal dynamics. To better capture multi-scale temporal features, motion representations are further processed by the Spiking Transformer module (SpikeFormer), which leverages self-attention and a multi-stage timestep shrinkage strategy to efficiently model long-range dependencies. Semantic and motion features from both streams are fused via residual cross-attention in the Cross-Modal Reasoning Module (CRM), resulting in a comprehensive representation. Additionally, a dynamic threshold block adjusts the excitability of spiking neurons based on motion entropy and semantic context richness. This cohesive architecture allows MDST++ to adaptively focus on meaningful information across both modalities, achieving robust generalization on unseen classes.

\subsection{Contextual Semantic Information Modeling}
\subsubsection{Audio and visual encoder} The pre-trained SeLaVi \cite{CV10asano2020labelling} is used to extract robust and discriminative audio and visual features. The audio and visual encoders denoted as $E_{aud}$ and $E_{vis}$, respectively, are designed to further explore the semantic information of different modalities. The outputs of the audio and visual encoders are:
$
\boldsymbol{a}^{t} = E_{aud}(\boldsymbol{\mathcal{X}}_{a}) \quad \text{and} \quad \boldsymbol{v}^{t} = E_{vis}(\boldsymbol{\mathcal{X}}_{v}),
$
where $\boldsymbol{a}^{t} \in \mathbb{R}^{D_{a} \times D_{emb}}$ and $\boldsymbol{v}^{t} \in \mathbb{R}^{D_{v} \times D_{emb}}$. Each encoder consists of a sequence of two linear layers $f_{1}^{m}$ and $f_{2}^{m}$ for $m \in (\boldsymbol{a}^{t}, \boldsymbol{v}^{t})$. Specifically:
$
f_{1}^{m}: \mathbb{R}^{D_{m} \times T_{in}} \rightarrow \mathbb{R}^{D_{m} \times T_{hid}} \quad \text{and} \quad f_{2}^{m}: \mathbb{R}^{D_{m} \times T_{hid}} \rightarrow \mathbb{R}^{D_{m} \times T_{emb}}.
$
Each linear layer is followed by batch normalization, a ReLU activation function, and dropout with a rate of $d_{enc}$. Therefore, the $E_{aud}$ and $E_{vis}$ encoders can further explore the semantic information of each modality, enhancing the representation of audio and visual features more discriminatively.

\subsubsection{Recurrent Joint Learning Unit. (RJLU)} To improve the efficiency of audio-visual joint learning and incorporate global temporal information into extracted contextual semantic features, we introduce the Recurrent Joint Learning Unit (RJLU), depicted in the bottom-left corner of Fig. \ref{fig:2}. This unit extracts global joint knowledge from audio and visual features and iteratively updates it using a recurrent structure.

Firstly, audio features $\boldsymbol{a}^{t}$ and visual features $\boldsymbol{v}^{t}$ at time step $t$ are combined with the joint knowledge $\boldsymbol{h}^{t-1}$ from the previous time step $t-1$, as described by:
\begin{equation}
    \begin{aligned}
        \boldsymbol{C}_{a}^{t} &= \mathrm{CA}(\mathrm{AP}(\boldsymbol{a}^{t}), \mathrm{AP}(\boldsymbol{h}^{t})), \\
        \boldsymbol{C}_{v}^{t} &= \mathrm{CA}(\mathrm{AP}(\boldsymbol{v}^{t}), \mathrm{AP}(\boldsymbol{h}^{t})),
    \end{aligned}
\end{equation}
where $\mathrm{AP}(\cdot)$ denotes the average pooling function and $\mathrm{CA}(\cdot)$ signifies the cross-attention function. To enhance efficient joint learning, joint knowledge is a buffer linking audio and visual features. The self-attention function further identifies the intrinsic connections between these linked features. The output of the RJLU is defined as:
\begin{equation}
    \begin{aligned}
        \boldsymbol{C}_{ahv}^{t} &= \mathrm{CAT}(\boldsymbol{a}^{t}, \boldsymbol{h}^{t-1}, \boldsymbol{v}^{t}), \\
        \boldsymbol{S}_{ahv}^{t} &= \mathrm{MLP}(\mathrm{LN}(\mathrm{SA}(\boldsymbol{C}_{ahv}^{t}))) + \mathrm{SA}(\boldsymbol{C}_{ahv}^{t}),
    \end{aligned}
\end{equation}
where $\mathrm{CAT}(\cdot)$ represents the concatenation operation, $\mathrm{SA}(\cdot)$ represents the self-attention function, $\mathrm{LN}(\cdot)$ represents the layer normalization, and $\mathrm{MLP}(\cdot)$ represents the multi-layer perceptron. The audio-visual joint features obtained at this time step will be used to update the previous joint knowledge. 

To capture accurate joint knowledge, we employ a hierarchical progressive update strategy, which connects $\boldsymbol{C}_{a}^{t}$, $\boldsymbol{C}_{v}^{t}$ and $\boldsymbol{S}_{ahv}^{t}$ in series. Overall, the joint knowledge $\boldsymbol{h}^{t}$ at time step $t$ could be written as:
		
\begin{equation}
	\begin{aligned}
		\boldsymbol{C}_{av}^{t}&=\boldsymbol{C}_{a}^{t}+\boldsymbol{C}_{v}^{t},
		\\\boldsymbol{h}^{t} &= \boldsymbol{C}_{av}^{t} * \boldsymbol{h}^{t-1}+ (1-\boldsymbol{C}_{av}^{t})*\boldsymbol{S}_{ahv}^{t}.
	\end{aligned}
\end{equation}

\subsection{Motion Information Modeling (MIM)}
% \subsubsection{Event Generation Model (EGM)} 
% The event camera introduces a novel imaging approach, distinct from traditional cameras. Unlike conventional cameras that capture images at regular intervals, the event camera detects brightness changes within a scene and generates an asynchronous, high-speed event stream. Each event signals a brightness change, either positive or negative, along with the precise timing and location in the image. An event is formally defined as:
% \begin{equation}
%     e_{k} = (x_{k}, y_{k}, t_{k}, p_{k}),
% \end{equation}
% where $(x_{k}, y_{k})$ is the pixel location when the event is triggered, $t_{k}$ represents the timestamp, and $p_{k} \in \{-1, 1\}$ represents the polarity of this event, indicating the direction of the change.

% In our event generation model, an event is triggered when there is a change in the magnitude $\Delta L(\boldsymbol{u}, t_{k})$ of the logarithm of brightness at a given pixel $\boldsymbol{u}$ and time $t_{k}$ that exceeds a predetermined threshold $C$, since the occurrence of the last event at the same pixel:
% \begin{equation}
%     \Delta L(\boldsymbol{u}, t_{k}) = L(\boldsymbol{u}, t_{k}) - L(\boldsymbol{u}, t_{k} - \Delta t_{k}) \geq p_{k} C,
% \end{equation}
% with $\Delta t_{k}$ denoting the time since the previous event. The ideal sensor generative model is described by Eq. (1) and (2). The visual motion features are extracted as $\boldsymbol{\mathcal{E}}{v} = {e{k}}{k=1}^{N}$, where $\boldsymbol{\mathcal{E}}{v}$ aligns with the dimension of $\boldsymbol{v}^{t}$, and positions without events are filled with 0.
\begin{figure}
	\centering
	\includegraphics[scale=0.45]{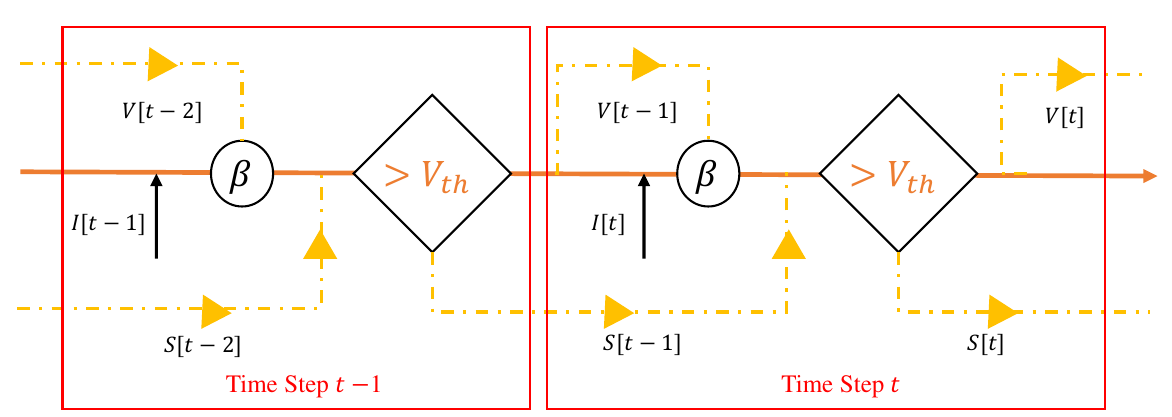}
	\caption{An illustration of a LIF neuron. The membrane potential \( V(t-1) \) and spike \( S(t-1) \) at time \( t-1 \) are derived from \( V(t-2) \) and \( S(t-2) \), and processed to produce \( U(t) \) and \( S(t) \) at time \( t \).}
	\label{fig:3}
\end{figure}
\subsubsection{Event Generation Model (EGM)}

The event camera significantly differs from traditional cameras because it uses a unique imaging technique. Instead of capturing images at fixed intervals, the event camera detects changes in brightness within a scene, generating an asynchronous, high-speed stream of events. Each event corresponds to a change in brightness, whether an increase or decrease and includes the timing and location within the image. An event is formally defined as:
\begin{equation}
    e_{k} = (x_{k}, y_{k}, t_{k}, p_{k}),
\end{equation}
where $(x_{k}, y_{k})$ is the pixel location when the event occurs, $t_{k}$ is the timestamp, and $p_{k} \in \{-1, 1\}$ indicates the polarity of the event, denoting the direction of the brightness change.

In our proposed EGM, an event is triggered when the change in the magnitude $\Delta L(\boldsymbol{u}, t_{k})$ of the logarithm of brightness at a specific pixel $\boldsymbol{u}$ and time $t_{k}$ exceeds a predefined threshold $C$ since the last event at the same pixel.:
\begin{equation}
    \Delta L(\boldsymbol{u}, t_{k}) = L(\boldsymbol{u}, t_{k}) - L(\boldsymbol{u}, t_{k} - \Delta t_{k}) \geq p_{k} C,
\end{equation}
with $\Delta t_{k}$ denotes the preceding event time. The ideal sensor generative model is described by Eq. (4) and Eq. (5). The visual motion features are extracted as $\boldsymbol{\mathcal{E}}_{v} = \{e_{k}\}_{k=1}^{N}$, where $\boldsymbol{\mathcal{E}}_{v}$ corresponds to the dimension of $\boldsymbol{v}^{t}$, and positions without events are filled with 0.

% \subsubsection{Discrepancy Analysis Block (DAB)}
% Audio features possess inherent temporal characteristics, yet defining their motion nature is challenging. We propose modeling the motion nature of audio features using visual event data, which we define as the difference in audio features. Formally, the discrepancy matrix is calculated as follows:
% \begin{equation}
%     \boldsymbol{\mathcal{E}}_{a} = \boldsymbol{a}^{t} + \left(1 - \exp\left(-\left\| \frac{\boldsymbol{\mathcal{E}}_{v} - \boldsymbol{a}^{t}}{\beta_{a}} \right\|^{2}_{2}\right)\right),
% \end{equation}
% where $\beta{a}$ is a learnable factor. Using the discrepancy matrix, SNNs can simultaneously extract both the motion and temporal characteristics of the audio features. This enables our model to capture a more comprehensive representation of the audio features.

\subsubsection{Discrepancy Analysis Block (DAB)}
Defining the motion nature of audio features is a complex task. We utilize visual event data to model the motion nature of audio features, which we define as the difference in audio features. The discrepancy matrix is formally computed as follows:
\begin{equation}
    \boldsymbol{\mathcal{E}}_{a} = \boldsymbol{a}^{t} + \left(1 - \exp\left(-\left\| \frac{\boldsymbol{\mathcal{E}}_{v} - \boldsymbol{a}^{t}}{\beta_{a}} \right\|^{2}_{2}\right)\right),
\end{equation}
where $\beta_a$ is a learnable parameter. Based on the discrepancy matrix, SNNs can concurrently explore the interrelationships between the motion and temporal natures of the audio features. This allows our model to capture a more comprehensive representation of the audio features.

\begin{figure*}
	\centering
	\includegraphics[scale=0.65]{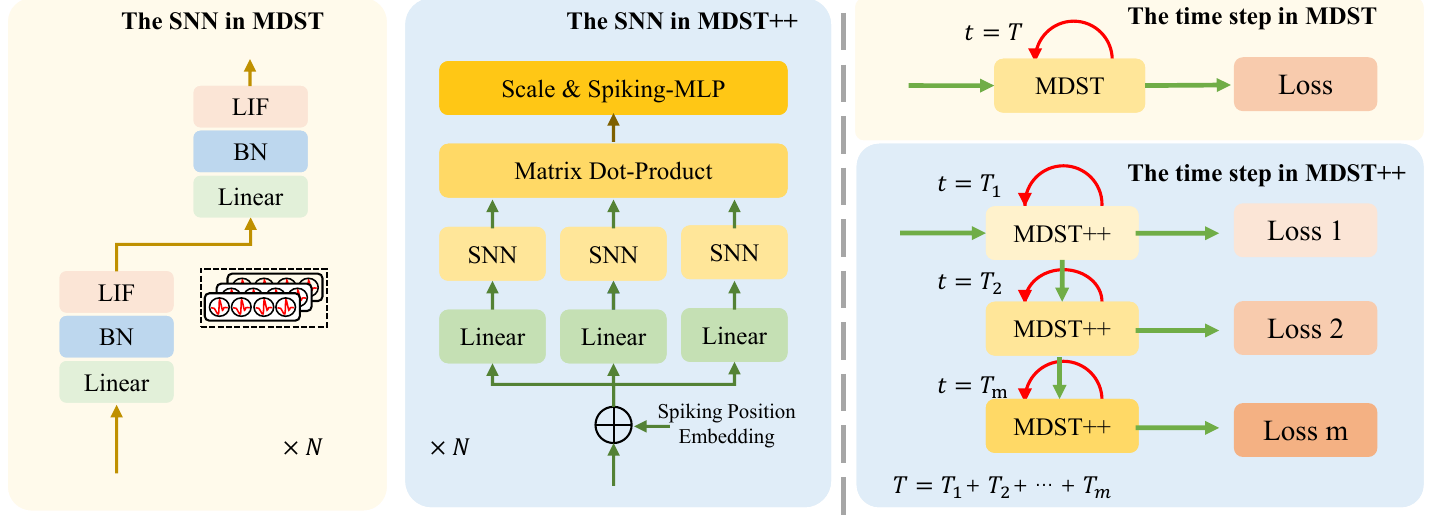}
% 	[height=8cm,width=18cm] [scale=0.78]
	\caption{Main differences between MDST and MDST++. In MDST, a simple SNN is built using three linear layers with LIF neurons. In MDST++, self-attention and Transformers are integrated to improve SNN's learning capability. Additionally, to leverage multi-scale time features, MDST++ divides the SNN into \( m \) stages with progressively compressed time steps (\( T_1 > T_2 > ... > T_m \)). The outputs at different time steps are used to calculate losses and train the model.}
	\label{fig:6}
\end{figure*}

\subsection{Spiking Transformer}
As depicted in Fig. \ref{fig:6}, MDST utilized a simple three-layer spiking MLP to process video events, assigning equal weights to outputs at all time steps. However, spiking MLPs often struggle to model temporal dependencies and asynchronous events naturally. MDST++ introduces an architecture that integrates SNNs with self-attention mechanisms and Transformers to address this limitation. This integration effectively captures long-range dependencies, enhances efficiency through parallel computation, and provides robust representational power, making it well-suited for handling complex patterns in sequential and event data. Additionally, we propose a multi-stage spiking timestep training strategy to leverage multi-scale temporal features. This strategy optimizes parameter updates without increasing inference costs and ensures the preservation of information by gradually reducing time steps, thereby preventing performance degradation.

SNNs are a type of neural network where communication between neurons occurs through brief pulses known as spikes. This method of spike-based communication closely resembles the event-driven nature of event cameras, thereby making SNNs particularly suitable for processing data from such devices. Our SNN architecture is composed of three linear SNN blocks, with each block consisting of a linear layer followed by a LIF neuron-based layer \cite{LIF/fnins.2016.00508}, as depicted in Fig. \ref{fig:3}. The Spiking Transformer (SpikeFormer) effectively incorporates self-attention mechanisms and Transformers, which enhance the ability of SNNs to detect complex patterns and relationships. The multi-scale time feature extraction splits the SNN into stages with gradually compressed time steps, allowing the model to capture both short-term and long-term dependencies effectively. Enhanced audio-visual fusion via cross-modal transformers ensures a comprehensive representation of multimodal data by merging scene contextual semantic and motion information. These design elements create a more robust and flexible model, capable of superior performance in various audio-visual tasks by effectively capturing and utilizing intricate temporal and semantic information. The overall spikeFormer architecture can be described as follows:
\begin{equation}
    \begin{aligned}
        \boldsymbol{Q} = \text{SNN}(\boldsymbol{\mathrm{W}}_{q}\boldsymbol{X}), 
        \boldsymbol{K} = \text{SNN}(\boldsymbol{\mathrm{W}}_{k}\boldsymbol{X}), 
        \boldsymbol{V} = \text{SNN}(\boldsymbol{\mathrm{W}}_{v}\boldsymbol{X}),
    \end{aligned}
\end{equation}
where $\text{SNN}(\cdot)$ represents the LIF layer, and $\boldsymbol{\mathrm{W}}_{q}$, $\boldsymbol{\mathrm{W}}_{k}$, and $\boldsymbol{\mathrm{W}}_{v}$ are learnable linear matrices. The spiking self-attention mechanism is then formulated as:
\begin{equation}
    \begin{aligned}
        \boldsymbol{G} &= \text{Scale}(\boldsymbol{QK}^{T}\boldsymbol{V}), \\
        \boldsymbol{S} &= \text{SNN}(\text{BN}(\text{Linear}(\boldsymbol{G}))),
    \end{aligned}
\end{equation}
where $ \text{Scale} (\cdot) $ acts as the scaling factor. By integrating self-attention and Transformers with SNNs, the Spiking Transformer can effectively process and interpret the complex temporal dynamics and relationships within audio-visual data, significantly enhancing the model's performance in zero-shot learning and related tasks.

\subsection{Multi-Stage Timestep Shrinkage}
As illustrated in Fig. \ref{fig:6}, SNNs with arbitrary architectures and layers are divided into $m$ stages. The timestep for each stage progressively decreases as the network deepens, i.e., $(T_1 > T_2 > ... > T_m)$. This strategy is based on the principle that the initial layers of the SNN require larger timesteps to extract features from the input. In comparison, subsequent layers utilize smaller timesteps to minimize inference latency. The gradual reduction in timestep, rather than an abrupt decrease, ensures the preservation of transmitted information and prevents performance degradation as the temporal resolution decreases.

Assuming the SNN is divided into $m$ stages, each requiring timesteps $\{T_1, T_2, ..., T_m\}$, the average timestep for inference with timestep shrinkage can be approximated as:
\begin{equation}
    T_{avg} = \frac{\sum_{i=1}^{m} m_{i} T_{i}}{\sum_{i=1}^{m} m_{i}},
\end{equation}
where $m_{i}$ represents the number of spiking units (such as spiking self-attention and spiking MLPs) in stage $i$. For SNNs without timestep shrinkage, the average timestep equals that of each stage. The strategy of using timestep shrinkage is based on several key principles:

1) \textbf{Efficiency in Feature Extraction}: The initial layers of the SNN require larger timesteps to extract meaningful features from the input data effectively. Larger timesteps allow these layers to capture more comprehensive information from the sparse event data provided by the event camera.

2) \textbf{Reduced Inference Latency}: Subsequent layers use smaller timesteps to reduce the inference latency. This reduction in timesteps enables faster processing of data, which is crucial for real-time applications such as robotics and autonomous driving.

3) \textbf{Gradual Reduction}: A gradual reduction in timesteps, rather than an abrupt decrease, ensures that transmitted information is retained effectively. This approach prevents performance degradation by maintaining the temporal resolution, which captures essential features at each network stage.

This strategy leverages the inherent strengths of SNNs in handling sparse event data. The dynamic adjustment of timesteps aligns with the goal of optimizing the network's efficiency and accuracy. By adopting this strategy, the network can better handle the temporal dynamics of input data, ultimately enhancing classification performance and mitigating the influence of background scene biases. The multi-stage timestep shrinkage approach ensures that each network layer is optimized for its specific role, leading to a more robust and efficient processing of complex audio-visual data.
\begin{figure}
	\centering
	\includegraphics[scale=0.65]{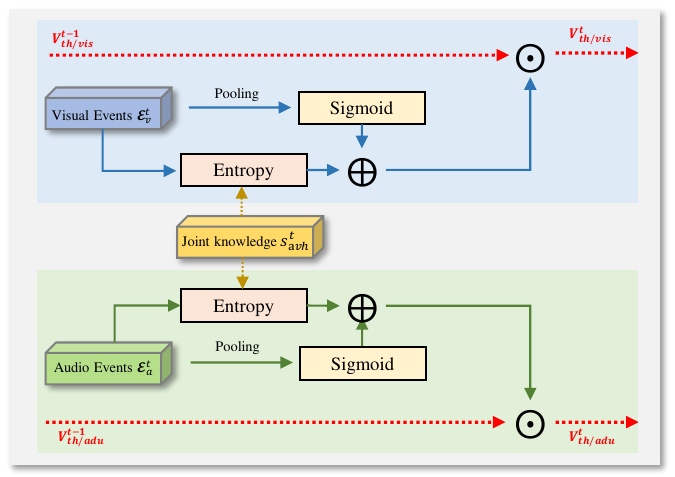}
% 	[height=8cm,width=18cm] [scale=0.78]
	\caption{The illustration of dynamic threshold block, which adaptively modifies the threshold $V_{th}^{t}$ of LIF neurons according to the statistics of scene contextual semantics and motion features.}
	\label{fig:4}
\end{figure}
\begin{figure}
	\centering
	\includegraphics[scale=0.65]{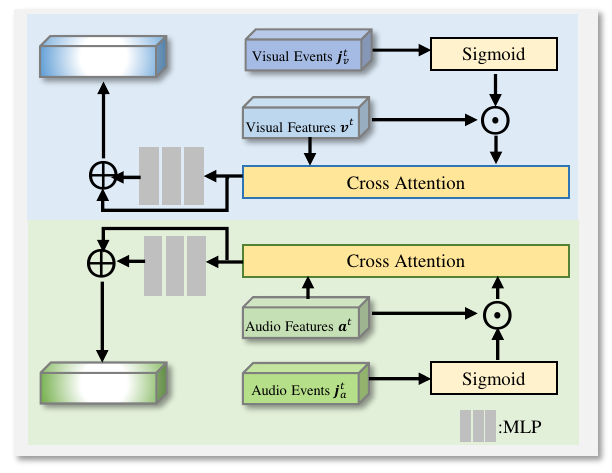}
% 	[height=8cm,width=18cm] [scale=0.78]
	\caption{The illustration of the fusion block effectively integrates the semantic and motion features between each modality.}
	\label{fig:5}
\end{figure}
\subsection{Dynamic Threshold Block}
Dynamic spiking thresholds have been proposed as a regulatory mechanism to prevent the excessive excitation and death of spiking neurons. In this block, we adaptively modify the threshold of LIF neurons according to the statistics of scene contextual semantics and motion features, as shown in Fig. \ref{fig:4}. 
To define the dynamic spiking thresholds required the global properties of contextual semantic features $\boldsymbol{\varphi}$ and the quantify of motion features $\boldsymbol{\omega}$. 
To define the dynamic spiking thresholds based on the semantic richness of the scene and the degree of motion dynamics. A contextual semantic score $\boldsymbol{\varphi}$ computed by applying a sigmoid function to the average-pooled joint semantic representation $\boldsymbol{S}_{ahv}^{t}$. This value reflects the global contextual semantics of the scene. A higher $\boldsymbol{\varphi}$ indicates richer semantic content, which leads to a lower neuron firing rate by reducing sensitivity. $\boldsymbol{\omega}$ quantifies motion information. It is calculated using an entropy-inspired formulation that combines the normalized semantic features with the event-based motion estimate $\boldsymbol{\mathcal{E}}_{v}^{t}$. A higher $\boldsymbol{\omega}$ indicates substantial motion variation, which leads to an increased spiking threshold to suppress noise. In contrast, a lower $\boldsymbol{\omega}$ results in a reduced threshold, enhancing the system's sensitivity to informative signals. Taking a visual pipeline as an example, the dynamic spiking threshold can be written as:
\begin{equation}
    \begin{aligned}
        V_{th}^{t} &= (\boldsymbol{\varphi} + \boldsymbol{\omega}) V_{th}^{t-1}, \\
        \boldsymbol{\varphi} &= \mathrm{Sigmoid}(\mathrm{AP}(\boldsymbol{S}_{ahv}^{t})), \\
        \boldsymbol{\omega} &= -\mathcal{N}(\boldsymbol{S}_{ahv}^{t}) \log\left(\frac{1}{\mathcal{N}(\boldsymbol{S}_{ahv}^{t})} + \boldsymbol{\mathcal{E}}_{v}^{t}\right),
    \end{aligned}
\end{equation}
where $V_{th}^{t}$ denotes the spiking threshold at time $t$, and $\mathcal{N}(\cdot)$ represents the normalization operation. 
\subsection{Cross-Modal Reasoning Module (CRM)}
The Cross-Modal Reasoning Module (CRM) aims to effectively combine the temporal and semantic features of audio-visual inputs from various modalities. To enhance the information exchange between audio and visual features, we employ a residual connection between two layers, followed by layer normalization. Fig. \ref{fig:5} depicts the fusion block's architecture. The outputs from the audio attention fusion block are defined as follows:
\begin{equation}
    \begin{aligned}
        \boldsymbol{R}_{a} &= \mathrm{CA}(\boldsymbol{a}^{t}, \boldsymbol{a}^{t} \ast \mathrm{Sigmoid}(\boldsymbol{j}_{a}^{t})), \\
        \boldsymbol{P}_{a} &= \mathrm{MLP}(\mathrm{LN}(\boldsymbol{R}_{a})) + \boldsymbol{R}_{a},
    \end{aligned}
\end{equation}
where $\boldsymbol{j}_{a}^{t}$ denotes the output of the SNN in the audio pipeline. The cross-attention function $\mathrm{CA}(\cdot)$ targets the most pertinent segments of the input features.

The cross-modal transformer delves deeper into the inherent relationships between fused semantic and temporal features and joint knowledge from different modalities. It amalgamates information from both audio and visual inputs to create a holistic joint audio-visual representation. The cross-modal transformer comprises multiple standard transformer layers and is described as follows:
\begin{equation}
    \begin{aligned}
        \boldsymbol{Z}_{av} &= \mathrm{MHCA}(\boldsymbol{P}_{v}, \boldsymbol{P}_{a}), \\
        \boldsymbol{F}_{av} &= \mathrm{MLP}(\mathrm{LN}(\boldsymbol{Z}_{av})) + \boldsymbol{Z}_{av},
    \end{aligned}
\end{equation}
where $\mathrm{MHCA}(\cdot)$ stands for the multi-head cross-attention mechanism, which efficiently integrates information from audio and visual modalities.

The primary objective is to predict the textually labeled class of the inputs. We employ projection and reconstruction layers to map the audio-visual joint embeddings into the textually labeled embedding space and reconstruct the projected features to restore the original information. This ensures the comparability of features from different modalities. The projection layer comprises two linear layers $f_{3}^{m}$ and $f_{4}^{m}$, each followed by batch normalization, a ReLU activation function, and dropout with a rate of $d_{proj}$. The functions of these layers are specified as:
$        f_{3}^{m}:  \mathbb{R}^{D_{m} \times T_{emb}} \rightarrow \mathbb{R}^{D_{m} \times T_{hid}}, 
        f_{4}^{m}: \mathbb{R}^{D_{m} \times T_{hid}} \rightarrow \mathbb{R}^{D_{m} \times T_{fin}}.$
The final audio-visual joint feature embeddings are obtained as follows:
\begin{equation}
    \begin{aligned}
        \boldsymbol{\mathcal{O}}_{av} &= \mathrm{AV}_{proj}(\boldsymbol{F}_{av}),
    \end{aligned}
\end{equation}
where $\mathrm{AV}_{proj}$ denotes the projection function. The final textual labeled embedding $\boldsymbol{\mathcal{O}}_{w}$ is derived by projecting the $j$-th class label embedding $\boldsymbol{w}_{j}$ using the word projection layer $W_{proj}$. The architecture of $W_{proj}$ is similar to $\mathrm{AV}_{proj}$ but has a different dropout rate $d_{wproj}$.

\subsection{Training Strategy}
The overall algorithm framework of MDST and MDST++ are illustrated in Algorithm \ref{algo1} and \ref{algo2}, respectively. The MDST model is trained on a single Nvidia V100S GPU, adhering to the method for extracting audio and visual embeddings per second as detailed in \cite{EC47mercea2022audio}. The model parameters are defined as follows: $T_{in} = 512$, $T_{hid} = 512$, $T_{proj} = 64$, and $T_{fin} = 300$. For the VGGSound, UCF, and ActivityNet datasets, the dropout rates are $d_{enc} = 0.20/0.25/0.10$, $d_{dec} = 0.25/0.20/0.15$, and $d_{wproj} = 0.1/0.1/0.1$, respectively. In the cross-modal transformer, we utilize a 8head with each head having a dimension of 64. The training process utilizes the Adam optimizer, and the MDST model is trained for 50 epochs at a learning rate of 0.0001. Our model's training is optimized to learn effective feature representations using the loss function $\mathcal{L}_{all}$, a combination of joint triplet loss $\mathcal{L}_{n}$, projection loss $\mathcal{L}_{p}$, and reconstruction loss $\mathcal{L}_{r}$.

\subsubsection{Joint Triplet Loss}

The joint triplet loss $\mathcal{L}_{n}$ plays a vital role in clustering the final audio-visual embeddings, ensuring both reasonableness and separation of results. This loss is formulated as:
\begin{equation}
    \mathcal{L}_{n} = [\gamma + \boldsymbol{\mathcal{O}}_{av}^{+} - \boldsymbol{\mathcal{O}}_{w}^{+}]_{+} + [\gamma + \boldsymbol{\mathcal{O}}_{av}^{-} - \boldsymbol{\mathcal{O}}_{w}^{+}]_{+},
\end{equation}
where $\gamma$ is the minimum margin separation between negative pairs of different modalities and the correctly matched audio-visual embeddings. Here, $\boldsymbol{\mathcal{O}}_{w}$ represents the textual embeddings, $\boldsymbol{\mathcal{O}}_{av}^{+}$ and $\boldsymbol{\mathcal{O}}_{av}^{-}$ denote the positive and negative examples, respectively, and $[x]_{+} \equiv \max(x, 0)$.

\subsubsection{Projection Loss}

The projection loss $\mathcal{L}_{p}$ aims to minimize the distance between the output joint embeddings from the projection layer and the corresponding textual embeddings.
\begin{equation}
    \mathcal{L}_{p} = \frac{1}{n} \sum_{i=1}^{n} \left\| \boldsymbol{\mathcal{O}}_{av} - \boldsymbol{\mathcal{O}}_{w} \right\|^2,
\end{equation}
where $n$ represents the number of samples.

% \subsubsection{Reconstruction Loss}
% The reconstruction loss $\mathcal{L}_{r}$ ensures the original data distribution is maintained while projecting audio-visual features into a shared embedding space. The reconstruction layer’s architecture mirrors that of the projection layer, and the reconstruction loss is expressed as:
% \begin{equation}
%     \mathcal{L}_{r} = \frac{1}{n} \sum_{i=1}^{n} \left\| \boldsymbol{\mathcal{O}}_{av}^{rec} - \boldsymbol{\mathcal{O}}_{w} \right\|^2,
% \end{equation}
% where $\boldsymbol{\mathcal{O}}_{av}^{rec}$ is the output of the reconstruction layer. The total loss function is formulated as follows:
% \begin{equation}
%     \mathcal{L}_{all} = \mathcal{L}_{n} + \mathcal{L}_{p} + \mathcal{L}_{r}.
% \end{equation}
% This combined loss function ensures that the model effectively learns to embed and reconstruct the audio-visual features, resulting in robust performance across various benchmarks.

\subsubsection{Reconstruction Loss}

The reconstruction loss $\mathcal{L}_{r}$ is designed to preserve the original data distribution while embedding features into a unified space. The architecture of the reconstruction layer is similar to that of the projection layer, and the reconstruction loss is defined as follows:
\begin{equation}
    \mathcal{L}_{r} = \frac{1}{n} \sum_{i=1}^{n} \left\| \boldsymbol{\mathcal{O}}_{av}^{rec} - \boldsymbol{\mathcal{O}}_{w} \right\|^2,
\end{equation}
where $\boldsymbol{\mathcal{O}}_{av}^{rec}$ represents the output of the reconstruction layer. The overall loss function is given by:
\begin{equation}
    \mathcal{L}_{all} = \mathcal{L}_{n} + \mathcal{L}_{p} + \mathcal{L}_{r}.
\end{equation}
This combined loss function ensures that the model learns to effectively embed and reconstruct the audio-visual features, achieving strong performance across various benchmarks.

% \section{Experiment}\label{experiment}
% In this study, we comprehensively evaluate our proposed model in both Zero-Shot Learning (ZSL) and Generalized Zero-Shot Learning (GZSL) scenarios. Following the methodology suggested by \cite{EC47mercea2022audio,EC71xian2018zero}, we calculate the mean class accuracy for all models to quantify their effectiveness in classification tasks. For the ZSL evaluation, we specifically analyze the test samples from the subset of unseen test classes. This allows us to assess the model's ability to generalize and accurately classify instances that belong to classes it has not encountered during training.

% In the GZSL evaluation, we assess the models on the entire test set, which includes both seen (S) and unseen (U) classes. This evaluation is crucial as it provides insights into the model's performance in more realistic scenarios where both known and novel classes are present. To measure the performance in GZSL, we calculate the harmonic mean (HM) of the accuracies on seen and unseen classes, defined as $ \mathrm{HM} = \frac{2 \mathrm{U}  \mathrm{S}}{\mathrm{U} + \mathrm{S}},$ where $\mathrm{U}$ and $\mathrm{S}$ represent the accuracies on unseen and seen classes, respectively. The harmonic mean provides a balanced measure that accounts for the trade-off between the accuracies on seen and unseen classes, ensuring a fair assessment of the model's generalization capabilities.
\begin{algorithm}[t]
\caption{Training framework for MDST}
\label{algo1}
\begin{algorithmic}[1]
\Require Input $\boldsymbol{\mathcal{X}}_{a,v} = (\boldsymbol{a}_{i}^{x}, \boldsymbol{v}_{i}^{x})$, label $\boldsymbol{\hat{Y}}_{a,v}$, timestep $T$
\Ensure Update network parameters $\mathbf{W}_{a,v}$
\State Initialize network parameters $\mathbf{W}_{a,v}$
\State $\boldsymbol{\mathcal{E}}_{a,v} \gets \text{EGM}(\boldsymbol{\mathcal{X}})$ \Comment{Convert input to events $\boldsymbol{\mathcal{E}}_{a,v}=(\boldsymbol{\mathcal{E}}_{a},\boldsymbol{\mathcal{E}}_{v})$}
\State $\boldsymbol{I}_{a,v} \gets \text{Encoder}(\boldsymbol{\mathcal{X}}_{a,v})$
\State $\boldsymbol{S}_{a,v} \gets \text{RJLU}(\boldsymbol{I}_{a,v})$ \Comment{Contextual inference}
\State $\boldsymbol{J}_{a,v} \gets \text{SNN}(\boldsymbol{\mathcal{E}}_{a,v})$ \Comment{Motion inference}
\State $\boldsymbol{F}_{a,v} \gets \text{Fusion}(\boldsymbol{S}_{a,v}, \boldsymbol{J}_{a,v})$ \Comment{Fuse features}
\State $\boldsymbol{Y}_{a,v} \gets \mathrm{AV}_{rec}(\boldsymbol{F}_{a,v})$ \Comment{The reconstruction output}
\State $\mathcal{L}_{all} \gets \mathcal{L}\left(\frac{1}{T} \sum_{t=1}^{T} \boldsymbol{Y}_t, \boldsymbol{\hat{Y}}_{a,v}\right)$ \Comment{Calculate loss}
\State Update parameters $\mathbf{W}_{a,v}$ by optimizing $\mathcal{L}_{all}$.
\end{algorithmic}
\end{algorithm}
\begin{algorithm}[t]
\caption{Training framework for MDST++}
\label{algo2}
\begin{algorithmic}[1]
\Require Input $\boldsymbol{\mathcal{X}}_{a,v} = (\boldsymbol{a}_{i}^{x}, \boldsymbol{v}_{i}^{x})$, label $\boldsymbol{\hat{Y}}_{a,v}$, number of stages $m$ of SNN, timesteps $\{T_1, T_2, \dots, T_m\}$
\Ensure Update network parameters $\mathbf{W}_{a,v}$
\State Initialize network parameters $\mathbf{W}_{a,v}$
\State $\boldsymbol{\mathcal{E}}_{a,v} \gets \text{EGM}(\boldsymbol{\mathcal{X}}_{a,v})$ \Comment{Convert input to events}
\State $\boldsymbol{I}_{a,v} \gets \text{Encoder}(\boldsymbol{\mathcal{X}}_{a,v})$
\For{$i = 1$ to $m-1$}
    \State $\boldsymbol{S}_{a,v,i} \gets \text{RJLU}(\boldsymbol{I}_{a,v})$ \Comment{Contextual inference}
    \State $\boldsymbol{J}_{a,v,i} \gets \text{SpikeFormer}_i(\boldsymbol{\mathcal{E}}_{a,v})$ \Comment{Motion inference}
    \State $\boldsymbol{F}_{a,v,i} \gets \text{Fusion}(\boldsymbol{S}_{a,v,i}, \boldsymbol{J}_{a,v,i})$ \Comment{Fuse features}
    \State $\boldsymbol{Y}_{a,v,i} \gets \mathrm{AV}_{rec}(\boldsymbol{F}_{a,v,i})$ 
    \State $\mathcal{L}_i \gets \mathcal{L}_i\left(\frac{1}{T_{i+1}} \sum_{t=1}^{T_{i+1}} \boldsymbol{Y}_{a,v,i,t}, \boldsymbol{\hat{Y}}_{a,v}\right)$ 
    \State $\boldsymbol{I}_{a,v} \gets$ Transform output and shrink timestep
\EndFor
\State $\boldsymbol{S}_{a,v,m} \gets \text{RJLU}(\boldsymbol{I}_{a,v})$
\State $\boldsymbol{J}_{a,v,m} \gets \text{SpikeFormer}_m(\boldsymbol{\mathcal{E}}_{a,v})$
\State $\boldsymbol{F}_{a,v,m} \gets \text{Fusion}(\boldsymbol{S}_{a,v,m}, \boldsymbol{J}_{a,v,m})$
\State $\boldsymbol{Y}_{a,v} \gets \mathrm{AV}_{rec}(\boldsymbol{F}_{a,v,m})$ \Comment{The reconstruction output}
\State $\mathcal{L}_m \gets \mathcal{L}_m\left(\frac{1}{T_m} \sum_{t=1}^{T_m} \boldsymbol{Y}_{a,v,m,t}, \boldsymbol{\hat{Y}}_{a,v}\right)$
\State $\mathcal{L}_{\text{total}} \gets \sum_{i=1}^{m} \mathcal{L}_i$ \Comment{Total loss}
\State Update parameters $\mathbf{W}_{a,v}$ by optimizing $\mathcal{L}_{\text{total}}$.
\end{algorithmic}
\end{algorithm}
\section{Experiments}\label{experiment}
% In the experimental section, we evaluate our model comprehensively and thoroughly by demonstrating the performance of MDST under two different testing settings: Zero-Shot Learning (ZSL) and Generalized Zero-Shot Learning (GZSL). Following  \cite{EC47mercea2022audio, EC71xian2018zero}, we measure the performance of all models on classification tasks in different test settings by calculating their average class accuracy.

In the ZSL evaluation, we test samples from the unseen class subset. This setting allows us to assess the model's generalization capability when faced with categories not encountered during training. This evaluation is crucial for understanding how the model can extrapolate knowledge to new classes. In the GZSL evaluation, we apply the model to the entire test set, including both seen (S) and unseen (U) classes. This evaluation demonstrates the model's performance in realistic scenarios with known and new classes. To measure performance in GZSL, we calculate the harmonic mean (HM) of the accuracies on seen and unseen classes, defined as $ \mathrm{HM} = \frac{2 \mathrm{U} \mathrm{S}}{\mathrm{U} + \mathrm{S}},$ where $\mathrm{U}$ and $\mathrm{S}$ represent the accuracies on unseen and seen classes, respectively. The harmonic mean offers a balanced measure that accounts for the trade-off between accuracy on seen and unseen classes, ensuring a fair assessment of the model's generalization capabilities.

\subsection{Dataset Statistic}
In this study, we utilize three benchmark datasets, ActivityNet, VGGSound, and UCF101, to conduct our experiments and evaluate the proposed models.

\subsubsection{ActivityNet} The ActivityNet \cite{CV26caba2015activitynet} dataset is a widely used benchmark dataset for video understanding, specifically designed to support action recognition tasks. It includes 200 different categories of daily activities, with over 20,000 video clips totaling more than 849 hours. Each video clip in the dataset is precisely annotated with activity categories and time ranges, providing detailed training and evaluation data. 

% Additionally, the ActivityNet dataset offers a set of standardized evaluation metrics, such as mean Average Precision (mAP), for comparing the performance of different models. The dataset is publicly available and has garnered extensive support from both the academic and industrial communities. The annual ActivityNet Challenge provides researchers with a platform to showcase new technologies and methods, driving advancements in the field of video analysis.

\subsubsection{UCF101} The UCF101 \cite{CV64soomro2012ucf101} dataset is a widely used video action recognition dataset, containing 101 action categories with a total of 13,320 video clips. The video content encompasses a variety of daily activities, sports, and entertainment behaviors, sourced from YouTube, with a resolution of 320x240. The number of videos per category varies, with categories including but not limited to fencing, playing piano, dancing, and swimming. 

% The dataset is designed with high diversity and complexity, featuring action scenes with different backgrounds, camera angles, and lighting conditions, which increases the challenge of the recognition task. UCF101 includes not only single-person actions but also multi-person interaction scenes, providing rich visual information and action samples. The UCF101 dataset is divided into three training and testing splits, each ensuring category balance to facilitate fair evaluation of algorithm performance. Researchers can use these splits for cross-validation to ensure the generalization capability of their models.

\subsubsection{VGGSound} The VGGSound \cite{CV19chen2020vggsound} dataset is a widely used multimodal audio-visual dataset containing over 200,000 video clips, covering 309 different sound event categories. Each video clip is 10 seconds long and sourced from YouTube, with a clear correspondence between each video's audio and visual content. 

% The categories in the VGGSound dataset encompass a broad range of everyday sound events, such as musical instrument performances, natural sounds, transportation noises, animal sounds, and more, providing rich training and testing data for tasks in multimodal learning, audio recognition, and video analysis. Additionally, the videos in the dataset are recorded under various environmental conditions, lighting, and perspectives, increasing the diversity and challenge of the dataset. This diversity helps to better simulate complex real-world scenarios, thereby promoting advancements in related research fields.
\begin{table*}

	\centering
	\begin{threeparttable}
		\renewcommand\arraystretch{1.5}
		\caption{The results comparison between MDST and state-of-the-art methods for audio-visual (G)ZSL.}
		\label{TAB1}
		\setlength{\tabcolsep}{5.5pt}{
\begin{tabular}{cccccclcccclcccc}
\hline \hline
\multirow{2}{*}{Type} &
  \multirow{2}{*}{Model} &
  \multicolumn{4}{c}{VGGSound-GZSL} &
   &
  \multicolumn{4}{c}{UCF-GZSL} &
   &
  \multicolumn{4}{c}{ActivityNet-GZSL} \\ \cline{3-6} \cline{8-11} \cline{13-16} 
 &
   &
  S &
  U &
  \textit{HM} &
  \textit{ZSL} &
   &
  S &
  U &
  \textit{HM} &
  \textit{ZSL} &
   &
  S &
  U &
  \textit{HM} &
  \textit{ZSL} \\ \hline
\multirow{2}{*}{ZSL} &
 SJE \cite{CV7akata2015evaluation} &
  48.33 &
  1.10 &
  2.15 &
  4.06 &
   &
  63.10 &
  16.77 &
  26.50 &
  18.93 &
   &
  4.61 &
  7.04 &
  5.57 &
  7.08 \\
 &
 DEVISE \cite{CV27frome2013devise} &
  36.22 &
  1.07 &
  2.08 &
  5.59 &
   &
  55.59 &
  14.94 &
  23.56 &
  16.09 &
   &
  3.45 &
  8.53 &
  4.91 &
  8.53 \\
 &
  APN \cite{CV85xu2020attribute} &
  7.48 &
  3.88 &
  5.11 &
  4.49 &
   &
  28.46 &
  16.16 &
  20.61 &
  16.44 &
   &
  9.84 &
  5.76 &
  7.27 &
  6.34 \\
 &
  VAEGAN \cite{CV80xian2019f} &
  12.77 &
  0.95 &
  1.77 &
  1.91 &
   &
  17.29 &
  8.47 &
  11.37 &
  11.11 &
   &
  4.36 &
  2.14 &
  2.87 &
  2.40 \\ \hline
\multirow{5}{*}{\begin{tabular}[c]{@{}c@{}}Audio-visual\\ ZSL\end{tabular}} &
  CJME \cite{EC55parida2020coordinatedCV56} &
  8.69 &
  4.78 &
  6.17 &
  5.16 &
   &
  26.04 &
  8.21 &
  12.48 &
  8.29 &
   &
  5.55 &
  4.75 &
  5.12 &
  5.84 \\
 &
  AVGZSLNet \cite{EC46mazumder2021avgzslnet} &
  18.05 &
  3.48 &
  5.83 &
  5.28 &
   &
  52.52 &
  10.90 &
  18.05 &
  13.65 &
   &
  8.93 &
  5.04 &
  6.44 &
  5.40 \\
 &
  AVCA \cite{EC47mercea2022audio} &
  14.90 &
  4.00 &
  6.31 &
  6.00 &
   &
  51.53 &
  18.43 &
  27.15 &
  20.01 &
   &
  24.86 &
  8.02 &
  12.13 &
  9.13 \\
 &
  TCaF \cite{TCAF22} &
  9.64 &
  5.91 &
  7.33 &
  6.06 &
   &
  58.60 &
  21.74 &
  31.72 &
  24.81 &
   &
  18.70 &
  7.50 &
  10.71 &
  7.91 \\ 
   &
  Hyper-alignment \cite{Hyper} &
  13.22 &
  5.01 &
  7.27 &
  6.14 &
   &
  57.28 &
  17.83 &
  27.19 &
  19.02 &
   &
  23.50 &
  8.47 &
  12.46 &
  9.83
 \\
 & AVMST \cite{li2023modality} &14.14 & 5.28& 7.68& 6.61&& 44.08 &22.63& 29.91& 28.19&& 17.75 &9.90 &12.71 &10.37\\
 
  \hline
 & 
  MDST \cite{MDST}&
  16.14 &
  5.97 &
  8.72 &
  7.13 &
   &
  48.79 &
  23.11 &
  31.36 &
  31.53 &
   &
  18.32 &
  10.55 &
  13.39 &
  12.55  
  \\ &
    MDST++ &
  18.72 &
  6.14 &
  \textbf{9.25} &
  \textbf{8.48} &
   &
  52.41 &
  24.49 &
  \textbf{33.38} &
  \textbf{33.81} &
   &
  19.93 &
  10.88 &
  \textbf{14.07} &
  \textbf{14.12} \\
  \hline \hline
\end{tabular}

}
\end{threeparttable}
\end{table*}
\begin{table}
	\centering
	\begin{threeparttable}
		\renewcommand\arraystretch{1.2}
		\caption{The statistics for (G)ZSL$^{cls}$ test setting.}
		\label{TAB3}
		\setlength{\tabcolsep}{6pt}{
\begin{tabular}{c|cccc|c}
\hline \hline
\multirow{2}{*}{Dataset} & \multicolumn{4}{c|}{\# Classes} & \#Videos \\ \cline{2-6} 
                         & all   & train     & val(U)   & test(U)   & test(U)     \\ \hline
VGGSound-GZSL$^{cls}$            & 271   & 138   & 69     & 64     & 3200     \\ 
UCF-GZSL$^{cls}$                 & 48    & 30    & 12     & 6      & 845      \\ 
ActivityNet-GZSL$^{cls}$         & 198   & 99    & 51     & 48     & 4052     \\ \hline \hline
\end{tabular}}
\end{threeparttable}
\end{table}

\subsection{Results Comparison}  
We demonstrate the superiority of the MDST framework by comparing it with recent methods, as shown in Table \ref{TAB1}. Our results across three benchmark datasets, VGGSound-GZSL, UCFGZSL, and ActivityNet-GZSL, highlight the significant improvements achieved by MDST.

% On the VGGSound-GZSL dataset, MDST achieves a Harmonic Mean (HM) of 8.72 and a Zero-Shot Learning (ZSL) performance of 7.13. Compared to the AVCA method, MDST shows a substantial improvement, with HM and ZSL scores increasing by 38.2\% and 18.8\%, respectively. MDST++ further enhances these results, achieving an HM of 9.25 and a ZSL performance of 8.48. These significant improvements can be attributed to MDST's effective decoupling of contextual semantic information from dynamic motion information, which mitigates background scene biases that often hinder classification accuracy.

For the UCFGZSL dataset, MDST achieves a Generalized Zero-Shot Learning (GZSL) performance of 31.36, slightly lower than TCaF's 31.72. We attribute TCaF's slight edge to its preprocessing steps that temporally align audio and visual features, enhancing its performance. However, MDST outperforms TCaF in the ZSL setting with a score of 29.34 compared to TCaF's 24.81. MDST++ further improves these results, achieving a GZSL performance of 33.38 and a ZSL performance of 33.81. This indicates that MDST++'s approach is particularly effective in scenarios with no training data for the classes, leveraging its superior handling of multi-scale temporal information to capture intricate temporal dynamics and improve overall performance.

On the ActivityNet-GZSL dataset, MDST significantly outperforms AVGZSLNet in GZSL performance, achieving 13.39 compared to AVGZSLNet's 6.44. This large margin showcases MDST's robustness and adaptability in complex real-world video datasets. For ZSL, MDST achieves a score of 11.94, considerably higher than CJME's 5.84. MDST++ further enhances this performance, achieving a GZSL score of 14.07 and a ZSL score of 14.12. This again underscores MDST's strength in handling unseen class data by focusing on motion cues and mitigating background biases, with MDST++ demonstrating an enhanced ability to capture and utilize multi-scale temporal information for superior classification results.

The MDST framework consistently demonstrate superiority across mainstream video datasets, especially in the ZSL test setting. These results validate the effectiveness of our dual-stream architecture, which decouples and processes contextual and motion information separately, thus enhancing the model's ability to generalize to unseen classes and improve overall classification accuracy. MDST++'s superior handling of multi-scale temporal information further solidifies its position as a leading approach in audio-visual zero-shot learning.

\begin{table*}
	\centering
	\begin{threeparttable}
		\renewcommand\arraystretch{1.5}
		\caption{The results comparison between MDST and state-of-the-art methods with different audio/video extracting networks.}
		\label{TAB4}
		\setlength{\tabcolsep}{5pt}{
\begin{tabular}{cccccclcccclcccc}
\hline \hline
\multirow{2}{*}{Type} &
  \multirow{2}{*}{Model} &
  \multicolumn{4}{c}{VGGSound-GZSL$^{cls}$} &
   &
  \multicolumn{4}{c}{UCF-GZSL$^{cls}$} &
   &
  \multicolumn{4}{c}{ActivityNet-GZSL$^{cls}$} \\ \cline{3-6} \cline{8-11} \cline{13-16} 
 &
   &
  S &
  U &
  \textit{HM $\uparrow$} &
  \textit{ZSL $\uparrow$} &
   &
  S &
  U &
  \textit{HM $\uparrow$} &
  \textit{ZSL $\uparrow$} &
   &
  S &
  U &
  \textit{HM $\uparrow$} &
  \textit{ZSL $\uparrow$} \\ \hline
\multirow{2}{*}{ZSL} &
 ALE \cite{ALE} &
  26.13 &
  1.72 &
  3.23 &
  4.97 &
   &
  45.42 &
  29.09 &
  35.47 &
  32.30 &
   &
  0.89 &
  6.16 &
  1.55 &
  6.16 \\
 &
  SJE \cite{CV7akata2015evaluation} &
  16.94 &
  2.72 &
  4.69 &
  3.22 &
   &
  19.39 &
  32.47 &
  24.28 &
  32.47 &
   &
  37.92 &
  1.22 &
  2.35 &
  4.35 \\
 &
 DEVISE \cite{CV27frome2013devise} &
  29.96 &
  1.94 &
  3.64 &
  4.72 &
   &
  29.58 &
  34.80 &
  31.98 &
  35.48 &
   &
  0.17 &
  5.84 &
  0.33 &
  5.84 \\
 &
  APN \cite{CV85xu2020attribute} &
  6.46 &
  6.13 &
  6.29 &
  6.50 &
   &
  13.54 &
  28.44 &
  18.35 &
  29.69 &
   &
  3.79 &
  3.39 &
  3.58 &
  3.97 \\ \hline
\multirow{5}{*}{\begin{tabular}[c]{@{}c@{}}Audio-visual\\ ZSL\end{tabular}} &
  CJME \cite{cjme} &
  10.86 &
  2.22 &
  3.68 &
  3.72 &
   &
  33.89 &
  24.82 &
  28.65 &
  29.01 &
   &
  10.75 &
  5.55 &
  7.32 &
  6.29 \\
 &
  AVGZSLNet \cite{EC46mazumder2021avgzslnet} &
  15.02 &
  3.19 &
  5.26 &
  4.81 &
   &
  74.79 &
  24.15 &
  36.51 &
  31.51 &
   &
  13.70 &
  5.96 &
  8.30 &
  6.39 \\
 &
  AVCA \cite{EC47mercea2022audio} &
  12.63 &
  6.19 &
  8.31 &
  6.91 &
   &
  63.15 &
  30.72 &
  41.34 &
  37.72 &
   &
  16.77 &
  7.04 &
  9.92 &
  7.58 \\
 &
  TCaF \cite{TCAF22} &
  12.63 &
  6.72 &
  8.77 &
  7.41 &
   &
  67.14 &
  40.83 &
  50.78 &
  44.64 &
   &
  30.12 &
  7.65 &
  12.20 &
  7.96  \\ 
  &        
  Hyper-alignment \cite{Hyper} &
  12.50 &
  6.44 &
  8.50 &
  7.25 &
   &
  57.13 &
  33.86 &
   42.52 &
  39.80 &
   &
29.77 & 8.77 & 13.55 & 9.13
 \\

  \hline 
& MDST \cite{MDST}&
  11.32 &
  8.71 &
  9.85&
  8.91 &
   &
  62.43 &
  43.12 &
51.01&
  49.23 &
   &
  26.36 &
  8.12 &
  12.42 &
  8.23 \\
   
& MDST++ &
  12.21 &
  8.38 &
  \textbf{9.94}&
  \textbf{10.32} &
   &
  65.32 &
  44.79 &
\textbf{53.14}&
  \textbf{55.22} &
   &
  27.53 &
  10.72 &
  \textbf{15.43} &
  \textbf{13.12} \\
\hline \hline
\end{tabular}
}
\end{threeparttable}
\end{table*}
\begin{table}
	\centering
	\begin{threeparttable}
				\renewcommand\arraystretch{1.2}
		\caption{The effectiveness of SNN in processing event data.}
		\label{TAB2}
		\setlength{\tabcolsep}{9pt}{
			% Please add the following required packages to your document preamble:
			% \usepackage{multirow}
\begin{tabular}{ccccc}
\hline \hline
\multirow{2}{*}{Model}            & \multicolumn{4}{c}{UCF-GZSL}                             \\ \cline{2-5} 
                                  & S     & U              & \textit{HM}    & \textit{ZSL}   \\ \hline
MLP  & 52.28 & 14.98          &  23.29        & 13.65          \\
EGM+MLP  & 46.52 & 16.34          & 24.19          & 18.93      \\ 
SNN   & 45.29 & 19.46          & 27.23          & 26.65          \\ \hline
MDST(EGM+SNN)                              & 48.79 & 23.11 & \textbf{31.36} & \textbf{31.53} \\ \hline \hline
\end{tabular}
		}
	\end{threeparttable}
\end{table}
\subsection{Different Audio/Visual Backbones}
We integrate features from pre-trained networks in audio and video classification into our methodology. Specifically, we use C3D \cite{C3D} for visual feature extraction, which leverages the Sports1M \cite{Sports1M} dataset. This results in a 4096-dimensional visual feature vector. For audio, we use VGGish \cite{VGGish}, trained on the Youtube-8M \cite{youtube8m} dataset, producing a 128-dimensional audio feature vector. We average these features over time to create unified video representations.

To ensure there is no overlap with classes in Youtube-8M, we adjust the dataset splits in VGGSound-GZSL, UCF-GZSL, and ActivityNet-GZSL. These adjusted splits are named VGGSound-GZSL$^{cls}$, UCF-GZSL$^{cls}$, and ActivityNet-GZSL$^{cls}$. Details of these changes are provided in Table \ref{TAB3}.

Table \ref{TAB4} shows the performance of MDST compared to baseline models. MDST consistently outperforms competitors. For instance, in the VGGSound-GZSL$^{cls}$ dataset, MDST achieves a Harmonic Mean (HM) of 9.94\% and a Zero-Shot Learning (ZSL) accuracy of 10.32\%, surpassing the TCaF model's HM of 8.77\% and ZSL accuracy of 7.41\%. In the UCF-GZSL$^{cls}$ dataset, MDST++ achieves an HM of 53.14\% and a ZSL accuracy of 55.22\%, considerably higher than MDST's HM of 51.01\% and ZSL accuracy of 49.23\%.

MDST++ introduces self-attention mechanisms and Transformers, which capture long-range dependencies more effectively than the simpler architecture in V1. Additionally, MDST++ employs a multi-stage spiking timestep training strategy, leveraging multi-scale temporal features. This strategy optimizes parameter updates without increasing inference costs and ensures the preservation of information by gradually reducing time steps. These enhancements allow MDST++ to better handle complex patterns in sequential and event data, resulting in improved performance across datasets.

\subsection{Ablation Study}
\subsubsection{Superiorities of SNN in Processing Event Information}
% Table \ref{TAB2} illustrates the superiority of SNNs in processing event information. In our MIM branch, we use a combination of the Event Generation Model (EGM) and SNN. For comparison, we replaced the three-layer SNN with a three-layer MLP and tested various module combinations. We conducted experiments using only SNN, only MLP, and a combination of EGM and MLP, denoted as "SNN," "MLP," and "EGM+MLP," respectively. The results clearly show that EGM+SNN achieves the best performance in both Generalized Zero-Shot Learning (GZSL) and Zero-Shot Learning (ZSL). While MLP performs best on seen classes, it shows a significant gap in performance on unseen classes and ZSL compared to our model. This discrepancy is likely due to background scene bias, which makes it challenging for the model to focus on the fine-grained and crucial motion cues within the scene. Interestingly, the combination of EGM and MLP performs even worse in GZSL and ZSL than using SNN alone. This highlights that traditional MLPs are not suitable for processing highly sparse event information, underscoring the effectiveness of SNNs for this task.
Table \ref{TAB2} demonstrates the superior capability of SNNs in processing event information. In our MIM branch, we integrate the Event Generation Model (EGM) with SNN. To evaluate the effectiveness of our approach, we replaced the three-layer SNN with a three-layer MLP and tested different module combinations. We conducted experiments using only SNN, only MLP, and a combination of EGM and MLP, labeled as ``SNN," ``MLP," and ``EGM+MLP," respectively. The results clearly show that the EGM+SNN combination achieves the highest performance in both GZSL and ZSL.

Although the MLP performs best on seen classes, it significantly underperforms on unseen classes and in ZSL compared to our model. This performance gap is likely due to background scene bias, which makes it difficult for the model to focus on the subtle and crucial motion cues within the scene. Notably, the combination of EGM and MLP performs even worse in GZSL and ZSL than using SNN alone. This outcome highlights the inadequacy of traditional MLPs for processing highly sparse event information, underscoring the effectiveness of SNNs in this context.
\subsubsection{Influence of Unimodal and Multimodal Input}
We evaluate the performance of the MDST model against a single-modality input framework, as shown in Table \ref{TAB5}. This assessment is done by modifying the cross-modal transformer's input to a unimodal configuration and training each branch separately. Using the UCF-GZSL dataset, the visual branch achieves a superior GZSL performance (HM) of 23.12, surpassing the audio branch's 15.78. The visual branch scores 17.16 for ZSL performance, while the audio branch achieves 13.78. However, when training the MDST model with combined audio and visual inputs, the performance significantly exceeds that of single-modality inputs. On the UCF-GZSL dataset, the GZSL performance increases to 31.36, and the ZSL performance reaches 29.34. These results highlight the importance of combining audio and visual inputs for GZSL and ZSL tasks in video classification. Integrating multimodal data enables the model to capture a more comprehensive representation of the video content, thereby enhancing classification accuracy and robustness.

\begin{table}
	\centering
	\begin{threeparttable}
				\renewcommand\arraystretch{1.2}
		\caption{Ablation study of different components on UCF-GZSL.}
		\label{TAB5}
		\setlength{\tabcolsep}{12pt}{
			% Please add the following required packages to your document preamble:
			% \usepackage{multirow}
\begin{tabular}{ccccc}
\hline \hline
\multirow{2}{*}{Model}            & \multicolumn{4}{c}{UCF-GZSL}                             \\ \cline{2-5} 
                                  & S     & U              & \textit{HM}    & \textit{ZSL}   \\ \hline
Only audio & 6.14 & 4.63 & 15.78 & 13.78  \\
Only visual  & 5.13 & 5.16 & 23.12 & 17.16  \\
 \hline
W/o MIM   & 50.64 & 17.74          & 26.34          & 22.78          \\
W/o RJLU &38.79  & 20.11          & 26.44         & 23.14          \\
W/o CRM  & 44.72 & 21.92          & 29.44          & 27.67          \\ \hline
MDST                              & 48.79 & 23.11 & \textbf{31.36} & \textbf{31.53} \\ \hline \hline
\end{tabular}
		}
	\end{threeparttable}
\end{table}
\subsubsection{Effectiveness of MDST Components}
Table \ref{TAB5} showcases the impact of various components within our model. We evaluated different versions of the MDST by excluding the RJLU, motion information modeling, and cross-modal reasoning module, referred to as ``W/o RJLU", ``W/o MIM", and ``W/o CRM", respectively. The ablation study reveals that the performance of the MDST model declines whenever any of these components are omitted, highlighting the importance of each one. The most crucial component is the MIM. The MIM helps mitigate noise from background scene biases by converting images into events. The robust temporal features extracted by the SNN capture interactions between different modalities, significantly enhancing zero-shot classification performance. Notably, when the MIM is removed, there is a substantial performance drop in unseen classes, despite a slight improvement in performance in seen classes. This is due to inconsistencies in data distribution between the SNN and RJLU outputs. Future work will aim to better integrate features from these components.

\begin{table}
	\centering
	\begin{threeparttable}
				\renewcommand\arraystretch{1.2}
		\caption{The effectiveness of different loss items on UCF-GZSL.}
		\label{TAB6}
		\setlength{\tabcolsep}{12pt}{
			% Please add the following required packages to your document preamble:
			% \usepackage{multirow}
\begin{tabular}{ccccc}
\hline \hline
\multirow{2}{*}{Loss}            & \multicolumn{4}{c}{UCF-GZSL}                             \\ \cline{2-5} 
                                  & S     & U              & \textit{HM}    & \textit{ZSL}   \\ \hline
                                  
W/o $\mathcal{L}_{n}$+$\mathcal{L}_{r}$ &28.71  & 8.52          & 13.14          & 10.11 \\
W/o $\mathcal{L}_{n}$+$\mathcal{L}_{c}$  &34.17  &  10.14          & 15.64          & 13.19          \\
W/o $\mathcal{L}_{n}$  & 40.77 & 16.79          & 23.78          & 16.98          \\
W/o $\mathcal{L}_{r}$  & 42.78 & 16.96          & 24.29          & 19.82          \\
W/o $\mathcal{L}_{p}$  &43.17  &19.27           & 26.65          & 23.11           \\ \hline
MDST                             & 48.79 & 23.11 & \textbf{31.36} & \textbf{31.53} \\ \hline \hline
\end{tabular}
		}
	\end{threeparttable}
\end{table}
\subsubsection{Impact of Different Loss Items}
Table \ref{TAB6} shows the effects of different loss functions on model performance. Our experiments indicate that using the complete loss function results in the best HM and ZSL performance across the UCF-GZSL, VGGSound-GZSL, and ActivityNet-GZSL datasets. Specifically, for the UCF-GZSL dataset, excluding the loss function $\mathcal{L}_{n}$ during training resulted in the most significant decline, with HM and ZSL scores of 23.91 and 16.98, respectively. In contrast, using the complete loss function $\mathcal{L}_{all}$ achieved HM and ZSL scores of 31.36 and 29.34, respectively. These results highlight each loss function's crucial role in training the model. Therefore, incorporating the complete loss function in the training process is essential for achieving superior GZSL and ZSL performance with the MDST model.
\begin{table}
	\centering
	\begin{threeparttable}
				\renewcommand\arraystretch{1.2}
		\caption{Ablation study of different threshold in EGM.}
		\label{TAB7}
		\setlength{\tabcolsep}{10pt}{

\begin{tabular}{ccccc}
\hline \hline
\multirow{2}{*}{Events range}            & \multicolumn{4}{c}{UCF-GZSL}                             \\ \cline{2-5} 
                                  & S     & U              & \textit{HM}    & \textit{ZSL}   \\ \hline
                                  
$-8<x<8$ &28.71  & 8.52          & 13.14          & 10.11 \\
$-10<x<10$ &34.17  &  10.14          & 15.64          & 13.19          \\
$-12<x<12$  & 40.77 & 16.79          & 23.78          & 16.98          \\
$-15<x<15$ & 42.78 & 16.96          & 24.29          & 19.82          \\
\hline
MDST++                           &  52.41 &  24.49 &  \textbf{33.38} &  \textbf{33.81}  \\ \hline \hline
\end{tabular}
		}
	\end{threeparttable}
\end{table}
\subsubsection{Effectiveness of Dynamic Threshold}
\begin{figure}
	\centering
	\includegraphics[scale=0.25]{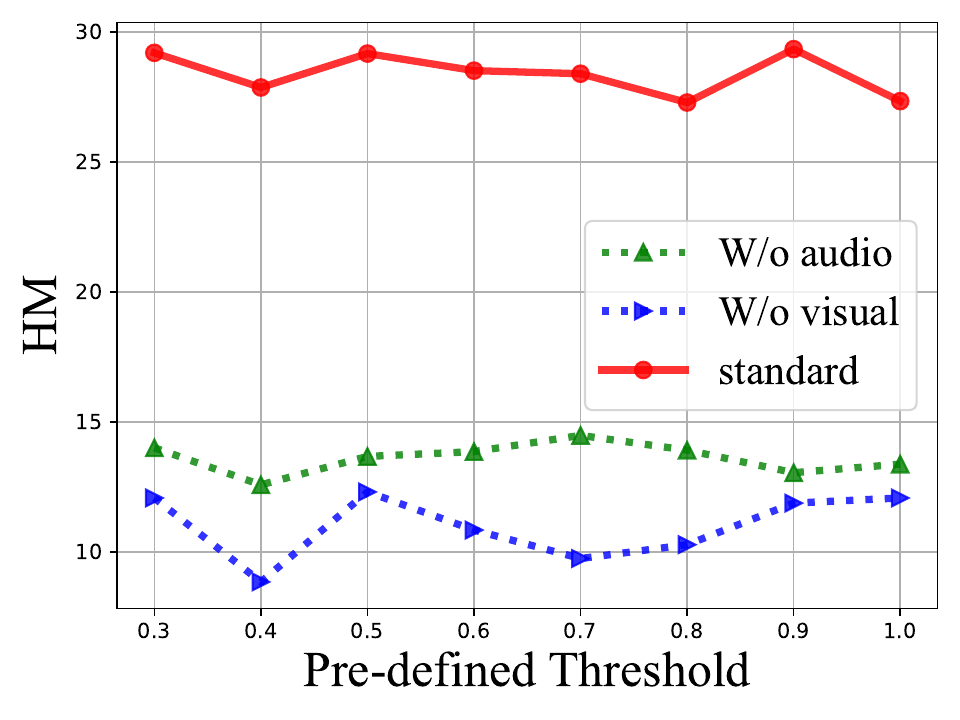}
	\includegraphics[scale=0.25]{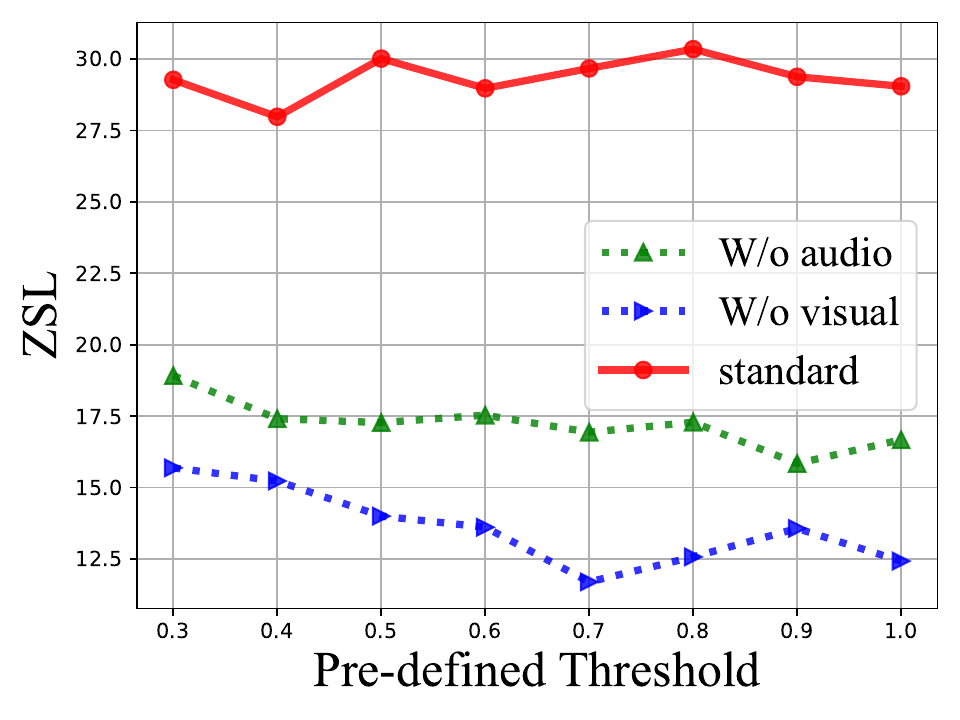}
	\caption{The comparison of different spiking thresholds on UCF-GZSL.}
	\label{fig:7}
\end{figure}

To demonstrate the benefits of using an adaptive spiking threshold and combining both audio and visual inputs, we present the training outcomes of our model with multimodal and unimodal inputs at different fixed spiking thresholds in Fig. \ref{fig:7}. The findings reveal significant performance variations in the model with different fixed spiking thresholds, indicating its sensitivity to these thresholds. Notably, the MDST model with a dynamic threshold consistently outperforms all methods using fixed spiking thresholds. Generally, models trained exclusively on visual inputs perform better than those trained solely on audio inputs. However, MDST integrates visual and audio modalities during training, exhibiting superior performance compared to unimodal models. This highlights that while visual features are crucial for audio-visual ZSL, incorporating audio features can further optimize the visual information.

\begin{table}
	\centering
	\begin{threeparttable}
				\renewcommand\arraystretch{1.2}
		\caption{Ablation study of each component between MDST and MDST++.}
		\label{TAB8}
		\setlength{\tabcolsep}{10pt}{
			% Please add the following required packages to your document preamble:
			% \usepackage{multirow}
\begin{tabular}{ccc}
\hline \hline
\multirow{2}{*}{Model}            & \multicolumn{2}{c}{UCF-GZSL}                             \\ \cline{2-3} 
         & \textit{HM}    & \textit{ZSL}   \\ \hline
SpikingMLP+Standard Timestep (MDST)           &  31.36        & 31.53          \\
SpikingMLP+Timestep Shrinkage            & 31.68          & 32.43      \\ 
SpikeFormer+Standard Timestep                              & 31.94 & 31.79 \\
SpikeFormer+Timestep Shrinkage (MDST++)                             & \textbf{33.38} & \textbf{33.81}
\\ \hline \hline
\end{tabular}
		}
	\end{threeparttable}
\end{table}

\subsubsection{Different EGM Thresholds}
\begin{figure}
	\centering
	\includegraphics[scale=0.25]{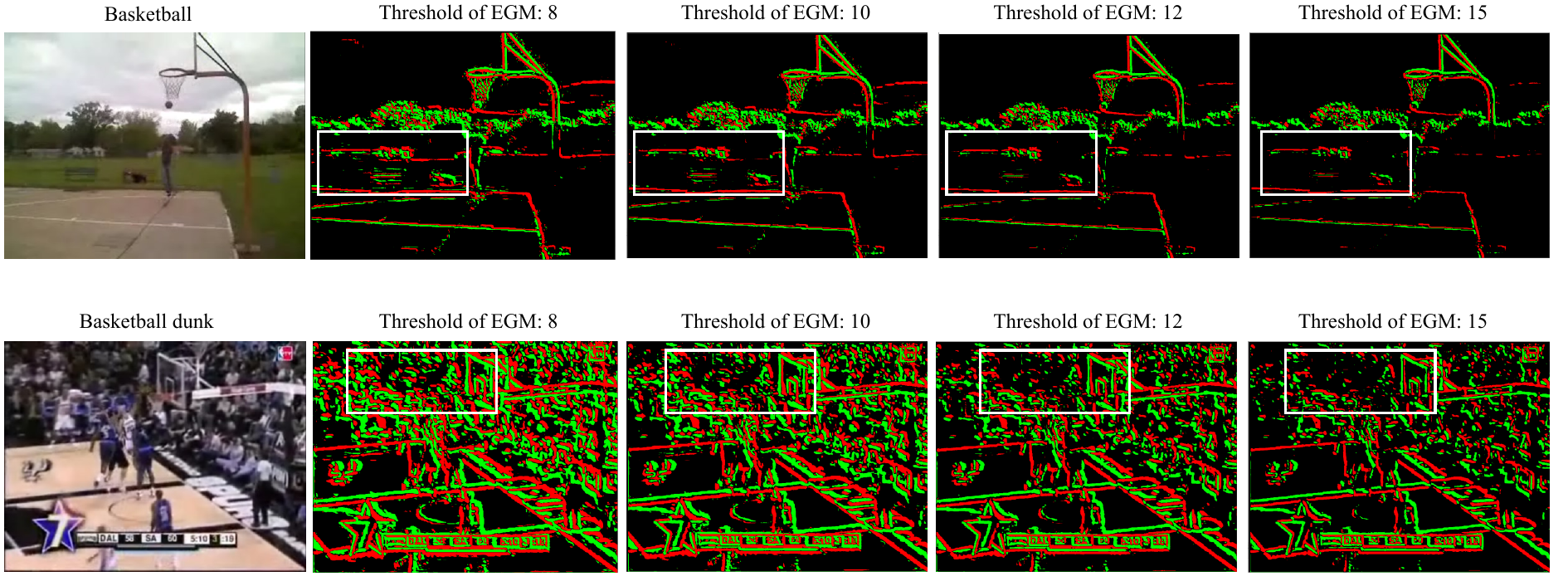}
% 	[height=8cm,width=18cm] [scale=0.78]
	\caption{The visualization of different EGM thresholds.}
	\label{fig:8}
\end{figure}

We validate the effects of different Event Generation Model (EGM) thresholds on our model. Fig. \ref{fig:8} illustrates how thresholds affect the generation of synthetic events across various scenarios. Different thresholds have minimal impact on overall event generation in scenarios with minimal motion information, as shown in the first row of Fig. \ref{fig:8}. However, in scenarios with more complex motion information, such as those in the second row, thresholds significantly impact the model. For instance, an event generated with a threshold of 8 in the white box contains substantially more information than an event generated with a threshold of 15, much of which comes from the background audience. It's important to note that setting a higher threshold doesn't necessarily result in better filtering of background motion noise. For example, in classes with similar semantic contexts, like ``Basketball" and ``Basketball Dunk," differences in motion information, including spectator reactions, player movements, and ball positions, are important for classification. 

Dynamic thresholds are crucial for SNNs when processing various event data because it's difficult to determine whether background motion information is useful for classification tasks. To filter out unexpected background noise, we adjust the SNN threshold using the cross-entropy of the event information as an indicator. By dynamically adjusting the SNN threshold, we can effectively filter out noise from the background audience and focus on the object's motion. We trained models using different predefined EGM thresholds and compared their performance, as shown in Table \ref{TAB7}. Our results demonstrate little difference in performance between different EGM thresholds. However, when comparing the performance of SNNs with different threshold settings, we found that the SNN threshold had a more significant impact on the model's performance. This emphasizes the importance of using dynamic thresholds to optimize event data processing and improve classification accuracy.


\subsubsection{Effectiveness of Each Component Between MDST and MDST++}
We present the superiority of SpikeFormer and its components in processing information in Table \ref{TAB8}. We compare the performance of MDST and MDST++ by adopting different models and timestep strategies. Specifically, we replace the SpikingMLP with SpikeFormer and test various combinations of modules. We conduct experiments using SpikingMLP and SpikeFormer with both standard and shrunk timesteps, denoted as ``SpikingMLP+Standard Timestep (MDST)," ``SpikingMLP+Timestep Shrinkage," ``SpikeFormer+Standard Timestep," and ``SpikeFormer+Timestep Shrinkage," respectively. It is evident that ``SpikeFormer+Timestep Shrinkage" achieves the best HM and ZSL performance. The models using only SpikingMLP perform worse on both HM and ZSL, especially with the standard timestep. The SpikeFormer model, even with the standard timestep, significantly outperforms the SpikingMLP models. This indicates that integrating SpikeFormer and timestep shrinkage is crucial for superior performance. SpikeFormer’s superior performance can be attributed to its ability to handle sparsity and temporal dynamics more effectively than SpikingMLP. The timestep shrinkage enhances the model’s ability to focus on fine-grained motion details while reducing background noise.

\subsubsection{Effectiveness of Different Average Timesteps}
\begin{figure}
	\centering
	\includegraphics[scale=0.28]{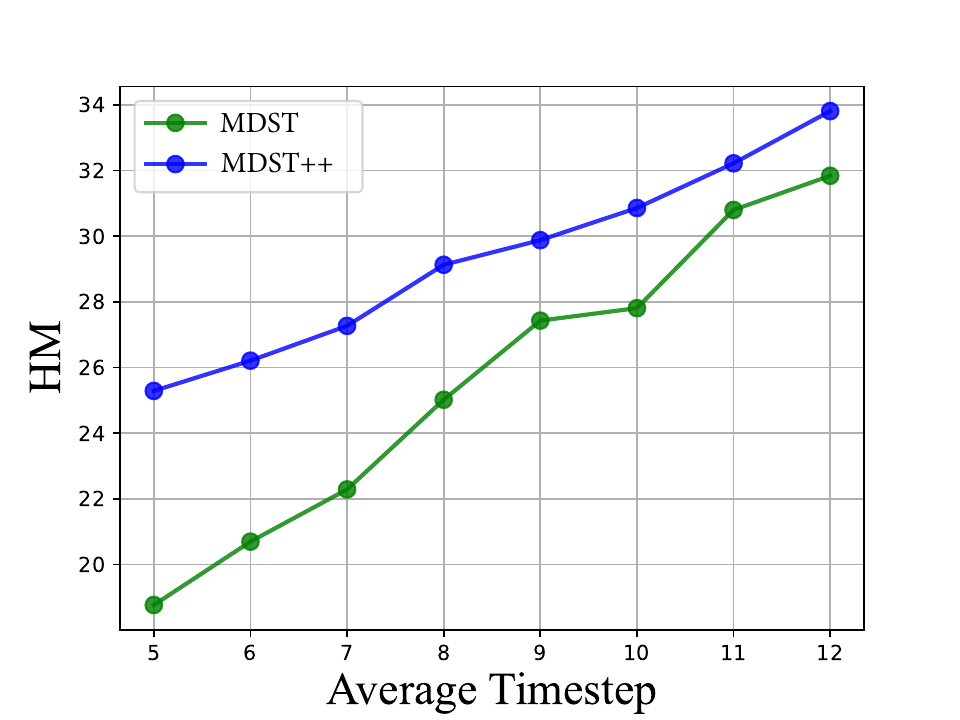}
	\includegraphics[scale=0.25]{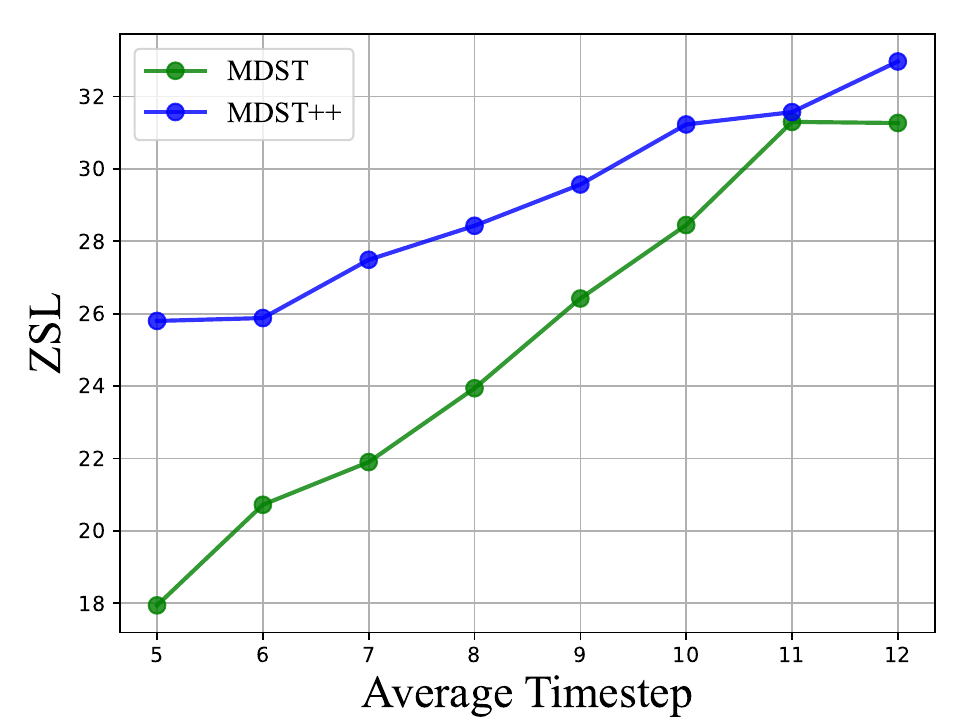}
	\caption{Ablation study on different average timestep of MDST and MDST++ in UCF-GZSL dataset.}
	\label{fig10}
\end{figure}

We assess the impact of average timestep on the performance of MDST and MDST++ using the UCF-GZSL dataset. The experiments are conducted with average timesteps ranging from 5 to 12; the results are illustrated in Fig. \ref{fig10}. As the average timestep increases, MDST++ consistently outperforms MDST. The performance difference between the two models becomes more pronounced with larger timesteps. Notably, MDST++ achieves higher performance at lower timesteps (e.g., 6, 7, 8) compared to MDST at higher timesteps (e.g., 10, 11, 12). This demonstrates MDST++'s ability to maintain high performance even with varying timestep settings, highlighting its robustness and efficiency in handling temporal information. The experiments reveal that MDST++'s approach to integrating SpikeFormer with dynamic timestep strategies significantly enhances its capability to process complex temporal patterns and mitigate background noise. 

\begin{table}
	\centering
	\begin{threeparttable}
		\renewcommand\arraystretch{1.2}
		\caption{The parameter comparison on UCF-GZSL dataset.}
		\label{revised-parameter}
		\setlength{\tabcolsep}{4pt}{
\begin{tabular}{c|cccccc}
\hline \hline
Model           & S     & U     & \textit{HM $\uparrow$}    & \textit{ZSL $\uparrow$}  &\#params &GFLOPS \\ \hline
AVCA \cite{EC47mercea2022audio}       &51.53 &  18.43 &  27.15 &  20.01 &1.69M	&2.36          \\
AVMST \cite{li2023modality}& 44.08 &  22.63 &  29.91 &  28.19 &6.32M	&5.12          \\
MDST \cite{MDST} &48.79 &  23.11 &  31.36 &  31.53 	&5.51M	&5.62 \\ \hline
MDST++           &\textbf{52.41} &  \textbf{24.49} &  \textbf{33.38} &  \textbf{33.81}  &6.21M &7.14\\ \hline \hline
\end{tabular}}
\end{threeparttable}
\end{table}
\subsection{Parameter Analysis}
We compare the parameter scale and computational complexity of MDST++ with several representative models, as summarized in Table \ref{revised-parameter}. Despite incorporating architectural enhancements like spiking self-attention and multi-stage timestep shrinkage, MDST++ maintains a relatively compact parameter size of 6.21M, comparable to AVMST of 6.32M and only slightly larger than the original MDST of 5.51M. Notably, this moderate increase leads to significant performance gains, with MDST++ achieving the highest HM of 33.38 and ZSL accuracy of 33.81 on the UCF-GZSL dataset. This suggests that the performance improvements are primarily due to more efficient modeling of temporal and semantic dynamics, rather than merely the increase in parameter size.

Despite these benefits, integrating SNNs and spiking transformers introduces new challenges in computational efficiency and practical deployment. To address these, MDST++ employs several mechanisms: (1) a multi-stage timestep shrinkage strategy reduces spike redundancy and inference latency by compressing timesteps as the network deepens; (2) dynamic threshold adjustment of LIF neurons modulates firing rates according to global motion and semantic entropy, enhancing sparsity and reducing unnecessary computation; and (3) a modular dual-stream design enables adaptive deployment, where either the semantic or motion stream can operate independently under limited resources. Furthermore, MDST++ is well-suited for neuromorphic hardware platforms (e.g., Intel Loihi or IBM TrueNorth), which naturally benefit from the event-driven and sparse characteristics of SNNs, offering potential for real-time, energy-efficient applications.

\subsection{Qualitative Results}
\begin{figure}
	\centering
	\includegraphics[scale=0.35]{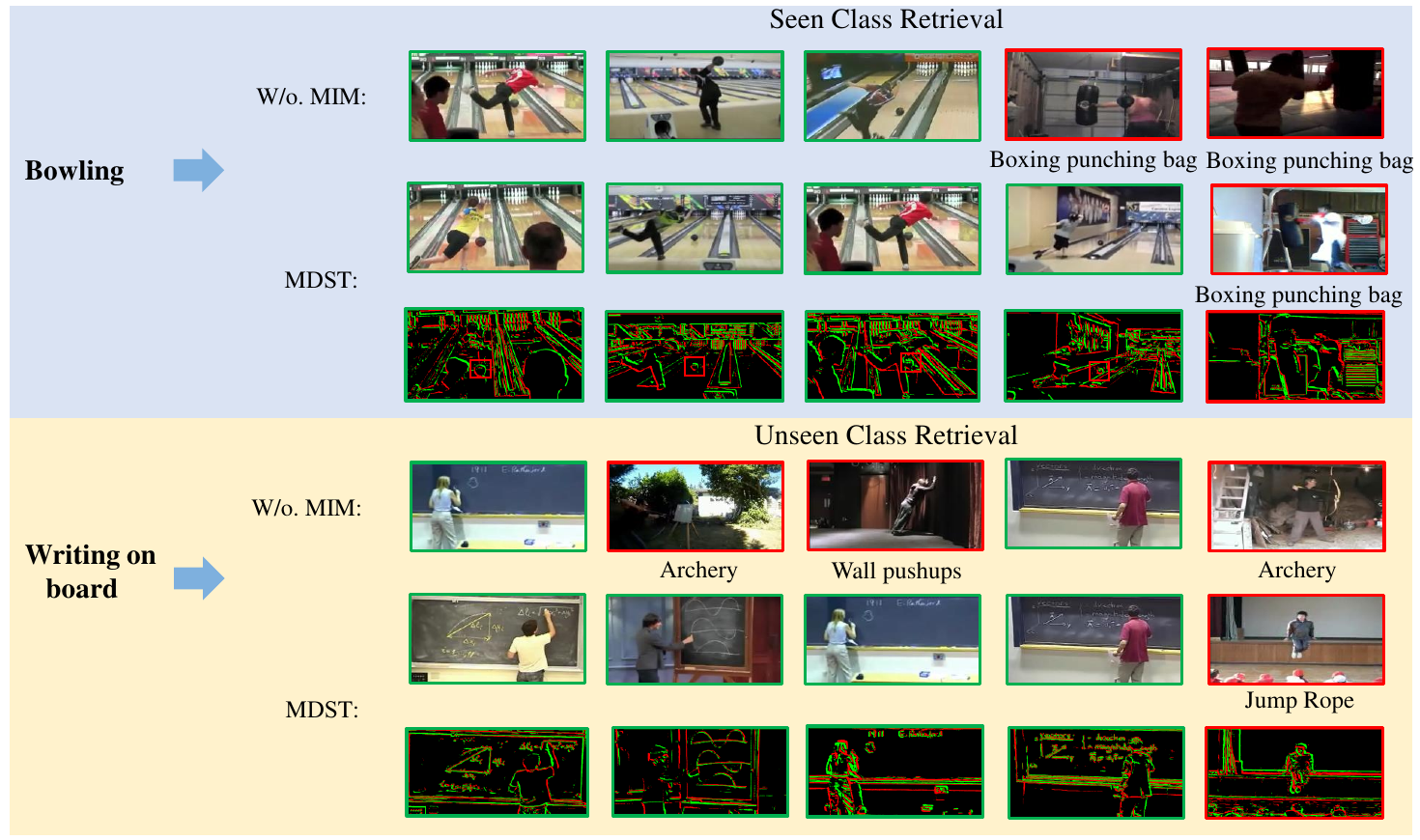}
	\caption{Qualitative comparison of ablation study. The correctly matched images are marked as green, and the mismatched images are marked as red. }
	\label{fig:9}
\end{figure}

To emphasize the importance of capturing motion information, we present qualitative results with and without the MIM branch visualization in Fig. \ref{fig:9}. For the ``Bowling" class in seen class retrieval, the absence of MIM leads to misclassifications like ``Boxing punching bag," showcasing the effects of scene bias. However, the model with MIM accurately identifies the bowling actions, correctly retrieving all ``Bowling" images. The MIM branch effectively removes background scene biases by converting images into events, leading to more precise classification results. This demonstrates MIM's critical role in capturing essential motion information.
\section{Conclusion}\label{conclusion}
In conclusion, we introduced the Motion-Decoupled Spiking Transformer framework to address the challenges of background scene bias and insufficient motion detail in audio-visual zero-shot learning. We effectively decoupled contextual semantic information from dynamic motion data by utilizing an event generation model to convert RGB images into events and leveraging spiking neural networks to process sparse event data. Our dual-stream architecture demonstrated superior performance in mitigating scene bias and enhancing motion information extraction. The recurrent joint learning unit facilitated efficient joint learning across multiple modalities, and integrating a discrepancy analysis block allowed for the effective modeling of audio motion features. We introduced the MDST++ to further improve the model's capability to capture long-range dependencies and multi-scale temporal features by combining SNNs with self-attention mechanisms. Extensive experiments on benchmark datasets validated the efficacy of our approach, with MDST and MDST++ consistently outperforming state-of-the-art methods in both zero-shot learning and generalized zero-shot learning scenarios. Our ablation studies confirmed the effectiveness of each key component in our framework, highlighting the robustness and adaptability of MDST++ in complex real-world video datasets.

%{\appendices
%\section*{Proof of the First Zonklar Equation}
%Appendix one text goes here.
% You can choose not to have a title for an appendix if you want by leaving the argument blank
%\section*{Proof of the Second Zonklar Equation}
%Appendix two text goes here.}

\bibliographystyle{IEEEtran}
% \bibliography{tpami}

\begin{IEEEbiography}[{\includegraphics[width=1in,height=1.25in,clip,keepaspectratio]{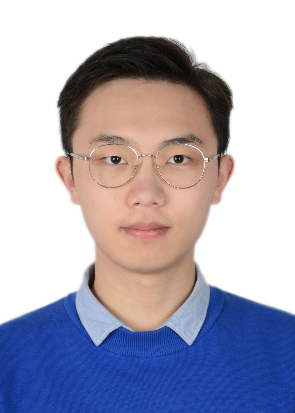}}]{Wenrui Li} received the B.S. degree from the School of Information and Software Engineering, University of Electronic Science and Technology of China (UESTC), Chengdu, China, in 2021. He is currently working toward the Ph.D. degree from the School of Computer Science, Harbin Institute of Technology (HIT), Harbin, China. His research interests include multimedia search, zero-shot learning, joint source-channel coding, and spiking neural network.
\end{IEEEbiography}

\begin{IEEEbiography}[{\includegraphics[width=1in,height=1.25in,clip,keepaspectratio]{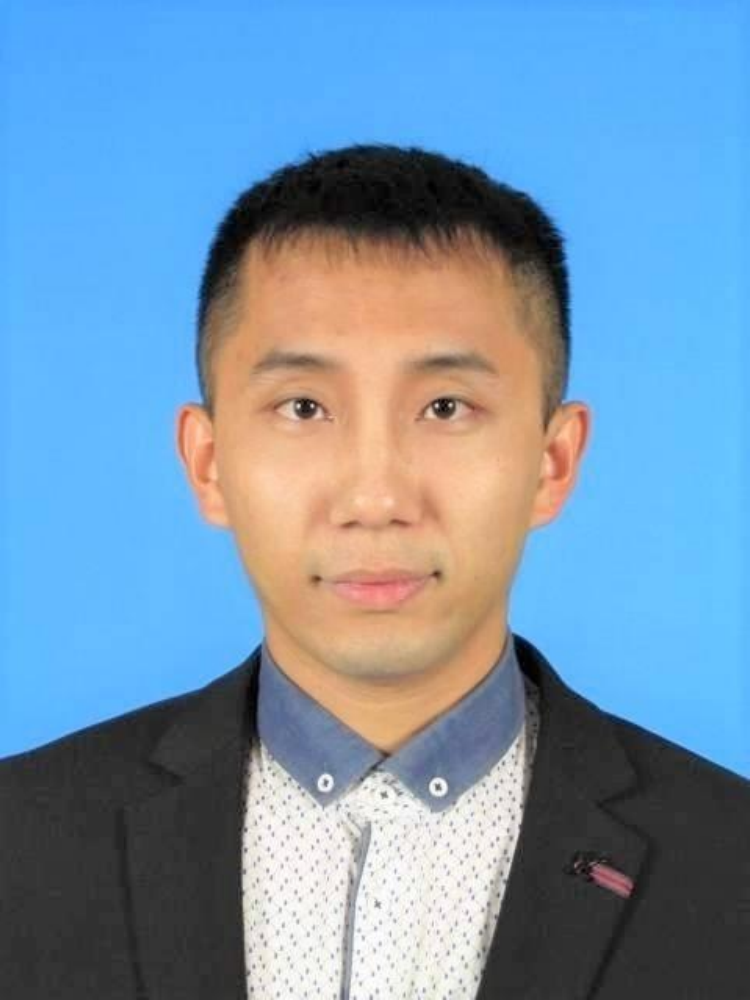}}]{Penghong Wang} received the M.S. degree in computer science and technology from Taiyuan University of Science and Technology, Taiyuan, China, in 2020. He is currently pursuing the Ph.D. degree with the School of Computer Science, Harbin Institute of Technology, Harbin, China. His main research interests include wireless sensor networks, joint source-channel coding, and computer vision.
\end{IEEEbiography}

\begin{IEEEbiography}[{\includegraphics[width=1in,height=1.25in,clip,keepaspectratio]{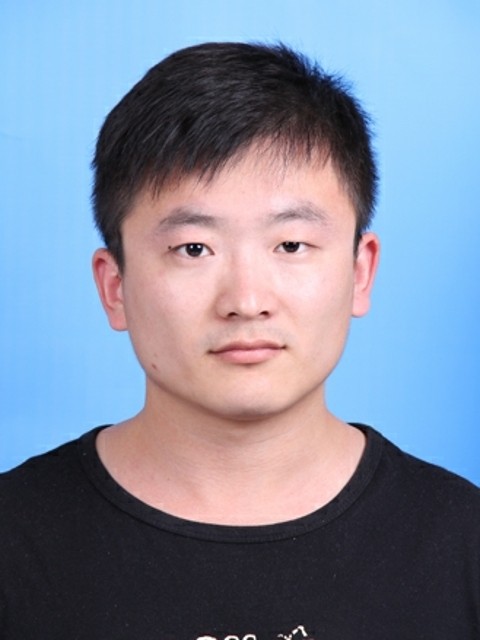}}]{Xingtao Wang} obtained his B.S. degree in Mathematics and Applied Mathematics, as well as his Ph.D. degree in Computer Science, from the Harbin Institute of Technology (HIT) in Harbin, China, in 2016 and 2022, respectively. In 2023, he served as an Assistant Research Fellow at the School of Artificial Intelligence, HIT, and currently holds the position of Associate Researcher. His research focuses on computer graphics, digital twins, and panoramic vision.
\end{IEEEbiography}

\begin{IEEEbiography}[{\includegraphics[width=1in,height=1.25in,clip,keepaspectratio]{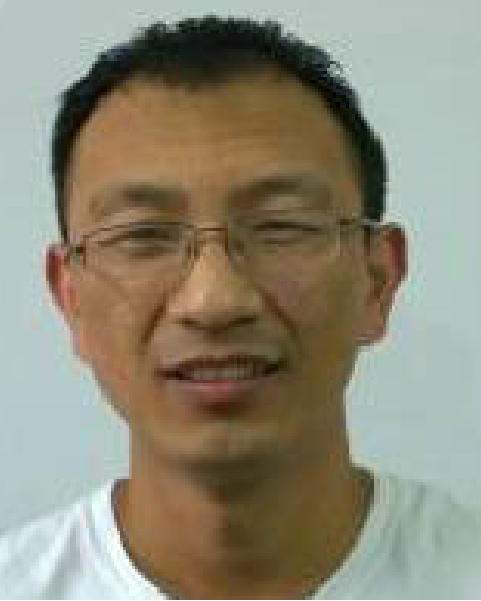}}]{Wangmeng Zuo} (Senior Member, IEEE) received the Ph.D. degree in computer application technology from Harbin Institute of Technology, Harbin, China, in 2007. He is currently a Professor with the Faculty of Computing, Harbin Institute of Technology.
He has published over 200 papers in top tier academic journals and conferences. His current research interests include low level vision, image/video generation, and multimodal understanding. He served as an Associate Editor for IEEE TRANSACTIONS
ON PATTERN ANALYSIS AND MACHINE INTELLIGENCE, IEEE TRANSACTIONS ON IMAGE PROCESSING, and SCIENCE CHINA Information Sciences.
\end{IEEEbiography}

\begin{IEEEbiography}[{\includegraphics[width=1in,height=1.25in,clip,keepaspectratio]{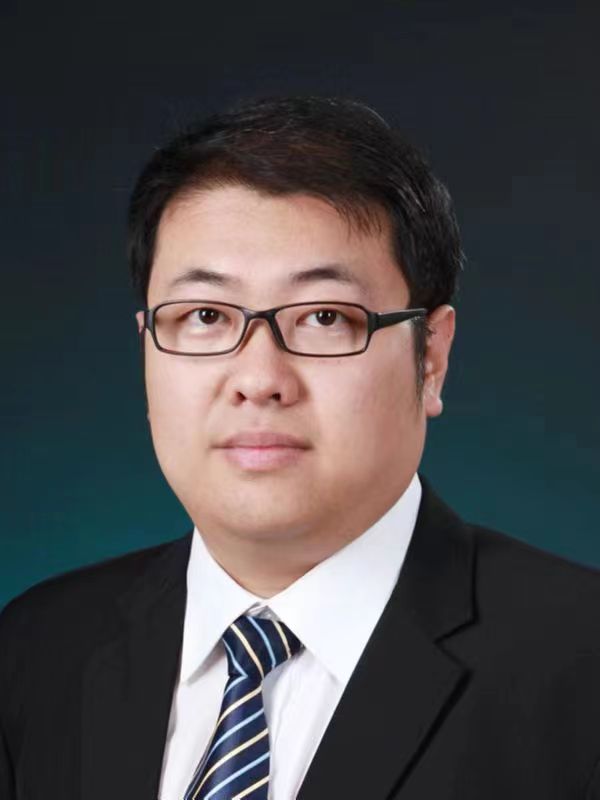}}]{Xiaopeng Fan} (Senior Member, IEEE) received the B.S. and M.S. degrees from the Harbin Institute of Technology (HIT), Harbin, China, in 2001 and 2003, respectively, and the Ph.D. degree from the Hong Kong University of Science and Technology, Hong Kong, in 2009. In 2009, he joined HIT, where he is currently a Professor. From 2003 to 2005, he was with Intel Corporation, China, as a Software Engineer. From 2011 to 2012, he was with Microsoft Research Asia, as a Visiting Researcher. From 2015 to 2016, he was with the Hong Kong University of Science and Technology, as a Research Assistant Professor. He has authored one book and more than 170 articles in refereed journals and conference proceedings. His research interests include video coding and transmission, image processing, and computer vision. He was the Program Chair of PCM2017, Chair of IEEE SGC2015, and Co-Chair of MCSN2015. He was an Associate Editor for IEEE 1857 Standard in 2012. He was the recipient of Outstanding Contributions to the Development of IEEE Standard 1857 by IEEE in 2013.
\end{IEEEbiography}

\begin{IEEEbiography}[{\includegraphics[width=1in,height=1.25in,clip,keepaspectratio]{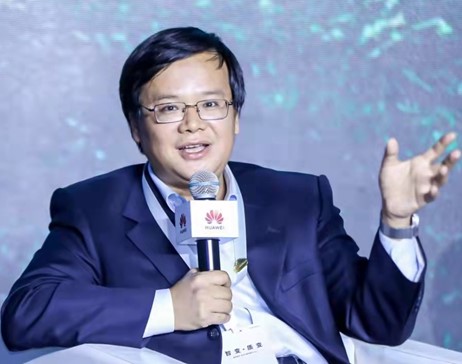}}]{Yonghong Tian} (Fellow, IEEE) is currently the Dean of the School of Electronics and Computer Engineering, a Boya Distinguished Professor with the School of
Computer Science, Peking University, China, and the Deputy Director of the Artificial Intelligence Research, Peng Cheng Laboratory, Shenzhen, China. He is the author or coauthor of over 350 technical papers in refereed journals and conferences. His
research interests include neuromorphic vision, distributed machine learning, and AI for science. He is a TPC Member of more than ten conferences, such as CVPR, ICCV, ACM KDD, AAAI, ACM MM, and ECCV. He is a Senior Member of CIE and CCF and a member of ACM. He was a recipient of the Chinese National Science Foundation for Distinguished Young Scholars in 2018, two National Science and Technology Awards, and three ministerial-level awards in China. He received the 2015 Best Paper Award for EURASIP Journal on Image and Video Processing, the Best Paper Award from IEEE BigMM 2018, and the 2022 IEEE SA Standards Medallion and SA Emerging Technology Award. He served as the TPC Co-Chair for BigMM 2015, the Technical Program Co-Chair for IEEE ICME 2015, IEEE ISM 2015, and IEEE
MIPR 2018/2019, and the General Co-Chair for IEEE MIPR 2020 and ICME 2021. He was/is an Associate Editor of IEEE TRANSACTIONS ON CIRCUITS AND SYSTEMS FOR VIDEO TECHNOLOGY from January 2018 to December 2021, IEEE TRANSACTIONS ON MULTIMEDIA from August 2014 to August
2018, IEEE Multimedia Magazine from January 2018 to August 2022, and IEEE ACCESS from January 2017 to December 2021. He co-initiated the IEEE International Conference on Multimedia Big Data (BigMM).

\end{IEEEbiography}

\end{document}